\pdfoutput=1

\documentclass[11pt]{article}

\usepackage{acl}                                                                           

\usepackage{times}
\usepackage{latexsym}
\usepackage{booktabs}
\usepackage{graphicx}
\usepackage{makecell}
\usepackage{multirow}
\usepackage{caption}
\usepackage{soul}
\usepackage{array} 
 \usepackage{amsmath}
\usepackage[T1]{fontenc}
\usepackage{caption}  
\usepackage{subcaption} 

\usepackage[utf8]{inputenc}

\usepackage{url}

\usepackage{microtype}

\usepackage{inconsolata}
\newcolumntype{L}[1]{>{\raggedright\arraybackslash}p{#1}}
\usepackage{graphicx}

\usepackage[x11names]{xcolor}
\newcommand{\cGreen}{Green1}
\definecolor{darkblue}{rgb}{0.0, 0.05, 0.7}

\hypersetup{
colorlinks   = true,
linkcolor    = darkblue,
citecolor    = darkblue,
urlcolor    =  darkblue
}
\title{Improving the Quality of Web-mined Parallel Corpora of Low-Resource Languages using Debiasing Heuristics}

\author{
 \textbf{Aloka Fernando\textsuperscript{1}},
 \textbf{Nisansa de Silva\textsuperscript{1}},
 \textbf{Menan Velayuthan\textsuperscript{1}},\\
 \textbf{Charitha Rathnayake\textsuperscript{1}}
 \textbf{Surangika Ranathunga \textsuperscript{2}}\\
\texttt{\{alokaf,NisansaDdS,velayuthan.22,charitha.18\}@cse.mrt.ac.lk}
\\
\texttt{s.ranathunga@massey.ac.nz}
\\
\textsuperscript{1}Dept. of Computer Science and Engineering, University of Moratuwa,10400, Sri Lanka\\
 \textsuperscript{2}School of Mathematical and Computational Sciences, Massey University, New Zealand
\\
 \small{
   \textbf{Correspondence:} \href{mailto:alokaf@cse.mrt.ac.lk}{alokaf@cse.mrt.ac.lk}
 }
}

\begin{document}
\maketitle

\begin{abstract}
Parallel Data Curation (PDC) techniques aim to filter out noisy parallel sentences from web-mined corpora. Ranking sentence pairs using similarity scores on sentence embeddings derived from Pre-trained Multilingual Language Models (multiPLMs) is the most common PDC technique. However, previous research has shown that the choice of the multiPLM significantly impacts the quality of the filtered parallel corpus, and the Neural Machine Translation (NMT) models trained using such data show a disparity across multiPLMs. This paper shows that this disparity is due to different multiPLMs being biased towards certain types of sentence pairs, which are treated as noise from an NMT point of view. We show that such noisy parallel sentences can be removed to a certain extent by employing a series of heuristics. The NMT models, trained using the curated corpus, lead to producing better results while minimizing the disparity across multiPLMs. We publicly release the source code and the curated datasets\footnote{\url{https://github.com/aloka-fernando/Heuristic_based_parallel_corpus_filtration}}.
\end{abstract}

\section{Introduction}\label{sec:introduction}
Parallel data mined from the web at scale is often considered an alternative to human-created data in training Neural Machine Translation (NMT) models~\cite{costa2022nllb,banon-etal-2020-paracrawl2}. CCAligned~\cite{el-kishky-etal-2020-ccaligned}, CCMatrix~\cite{schwenk-etal-2021-ccmatrix} and ParaCrawl~\cite{banon-etal-2020-paracrawl2} are examples of such web-mined corpora, which cover Low-Resource Languages (LRLs) as well. As summarised in Table~\ref{tab:flt_heuristics}, quality audits of these corpora reveal different types of noise. Consequently, training NMT models on such noisy parallel data leads to low-quality translations~\cite{khayrallah-koehn-2018-impact}.

\begin{table}[h]\centering
\scriptsize
\renewcommand{\arraystretch}{1.1}
\resizebox{0.48\textwidth}{!}{%
\begin{tabular}{lrrrrr}
\toprule
\textbf{Parallel Sentence Quality Category} &\textbf{A} &\textbf{B} &\textbf{C} &\textbf{D} &\textbf{E} \\
\midrule
Perfect translations & -& -& -&Y &Y\\
Near perfect translation & -& -& -& -&Y\\
Correct translation - Low quality & -&Y & -&Y &Y\\
Over/Under translation & -&Y &Y &Y&-\\
Misordered words &Y &Y &Y &Y &-\\
Spelling permutations & -&Y & -&Y &-\\
Untranslated Sentences &Y &Y &Y & -&Y\\
Short Sentences &Y & -&Y & -&Y\\
Mismatch numbers & -& Y& -& -&-\\
Machine Translated Sentences & -& -&Y &Y &-\\
Misaligned sentences &Y &Y &Y &Y &Y\\
Wrong Language &Y &Y &Y &Y &Y\\
Not a Language &Y &Y &Y &Y &Y\\
\hline
\end{tabular}}
\vspace{1mm}
\caption{Parallel sentence quality categories used in quality audits by~\citet{khayrallah-koehn-2018-impact} (\textbf{A}), \citet{bane-etal-2022-comparison} (\textbf{B}), \citet{herold-etal-2022-detecting} (\textbf{C}), 
\citet{kreutzer-etal-2022-quality} (\textbf{D}) and \citet{ranathunga-etal-2024-quality} (\textbf{E}).}\label{tab:flt_heuristics}
\end{table}

Parallel Data Curation (PDC) aims at extracting \textit{high-quality} parallel sentence pairs from noisy web-mined corpora. The importance of PDC for LRLs has been emphasised with the introduction of PDC shared tasks~\cite{sloto-etal-2023-findings,koehn-etal-2020-findings,koehn-etal-2019-findings}. Initiated by the work of ~\citet{chaudhary-etal-2019-low}, recent PDC techniques follow a \textit{scoring} and \textit{ranking} mechanism using embeddings obtained from a Multilingual Language Model (multiPLM). During the \textit{scoring} step, the semantic similarity is calculated between the source and target sentence embeddings for each sentence pair. Then the sentence pairs are \textit{ranked} in descending order of the similarity \textit{score}. Finally, a subset of the top-ranked sentence pairs is selected to train the NMT model. However, the quality of these top-ranked pairs depends on the chosen multiPLM~\citep{ranathunga-etal-2024-quality,moon2023doubts}.

To investigate the impact of using different multiPLMs on the PDC task, we conduct an initial analysis. We obtain embeddings from three multiPLMs: LASER3~\cite{heffernan-etal-2022-bitext}, XLM-R~\cite{conneau-etal-2020-unsupervised}, and LaBSE~\cite{feng-etal-2022-language}, calculate the semantic similarity between each parallel sentence pair in the CCMatrix~\cite{schwenk-etal-2021-ccmatrix} and CCAligned~\cite{el-kishky-etal-2020-ccaligned} datasets and rank them in descending order. Then we train NMT models (Section~\ref{sec:nmt_experiments}) using the top-ranked 100k sentence pairs from each corpus and report the ChrF++ scores. Experiments are carried out for English-Sinhala (En-Si), English-Tamil (En-Ta) and Sinhala-Tamil (Si-Ta) language-pairs. Sinhala, Tamil, and English belong to three distinct linguistic groups: Indo-Aryan, Dravidian, and Germanic (respectively). As shown in Figure~\ref{fig:disparity}, there is a significant difference, i.e.~a \textbf{\textit{disparity}} among these results, mainly for En-Si and En-Ta language pairs.

\begingroup
\setlength{\abovecaptionskip}{0pt}
\setlength{\belowcaptionskip}{0pt}
\begin{figure}[h]%
\centering
\includegraphics[trim={0cm 0cm 0cm 0cm}, clip, width=0.49\textwidth]{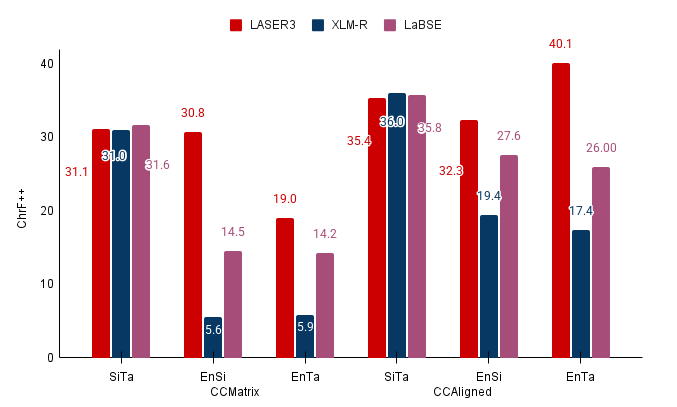}
\vspace{1mm}
\caption{Baseline NMT scores using top-ranked sentence pairs from CCMatrix and CCAligned corpora, with embeddings from LASER3, XLM-R, and LaBSE.}\label{fig:disparity}
\end{figure}
\endgroup

A random inspection of the En-Si top-ranked sentences reveals that different multiPLMs prioritise different sentence characteristics when ranking parallel sentences. For example, sentence pairs ranked top with XLM-R embeddings are mostly short and have high overlapping text such as numbers, acronyms, and URLs. LaBSE embeddings also result in sentences with numbers and date overlaps, while LASER3 embeddings result in relatively better full-length sentence pairs. This observation sheds light on the disparity in NMT results shown in Figure~\ref{fig:disparity} - the sentences ranked top when using XLM-R and LaBSE may have less linguistic content for NMT models to learn the translation. 

In order to carry out a more systematic evaluation, we randomly select 100 sentence pairs from the top 100K parallel sentences from the aforementioned ranked corpora and carried out a human evaluation.~\citet{ranathunga-etal-2024-quality}'s is the most comprehensive taxonomy available for this task. To better capture the noise observed during human evaluation, we extend this taxonomy (Section~\ref{sec:methodology}). Human evaluation results (Table~\ref{tab:HE_results}) indicate that the mean noise percentages are 98\%, 95\%, and 70\% when the corpora are ranked using LaBSE, XLM-R, and LASER3 embeddings, respectively. Further, the noise types in the categories of untranslated text (UN), short sentences (CS) and sentences with high-overlapping non-translatable text (CCN) could be observed, contributing to the reported percentages. Examples for these noise types are shown in Table~\ref{tab:examples_noisy_parallel_sentences} in Appendix~\ref{sec:AppendixHumanEvaluation}. We provide a detailed discussion of these results in Section~\ref{subsec:human_eval_results}. These findings suggest that inherent biases in multiPLMs lead to noisy parallel sentences being ranked highly, which in turn contributes to disparities across NMT models.

\begin{table*}[!htp]\centering
\renewcommand{\arraystretch}{1.1}
\resizebox{1.0\textwidth}{!}{
\begin{tabular}{llccc!{\vrule width 0.2pt}c!{\vrule width 0.2pt}cccccc!{\vrule width 0.2pt}c!{\vrule width 1.0pt}ccc!{\vrule width 0.2pt}c!{\vrule width 0.2pt}cccccc!{\vrule width 0.2pt}c!{\vrule width 1.0pt}ccc!{\vrule width 0.2pt}c!{\vrule width 0.2pt}cccccc!{\vrule width 0.2pt}c}
\toprule
\multirow{2}{*}{\textbf{}}&&\textbf{CC}&\textbf{CN}&\textbf{CB}&\textbf{\textit{C}}&\textbf{CS}&\textbf{CCN}&\textbf{UN}&\textbf{X}&\textbf{WL}&\textbf{NL}&\textbf{\textit{E}}&\textbf{CC}&\textbf{CN}&\textbf{CB}&\textbf{\textit{C}}&\textbf{CS}&\textbf{CCN}&\textbf{UN}&\textbf{X}&\textbf{WL}&\textbf{NL}&\textbf{\textit{E}}&\textbf{CC}&\textbf{CN}&\textbf{CB}&\textbf{\textit{C}}&\textbf{CS}&\textbf{CCN}&\textbf{UN}&\textbf{X}&\textbf{WL}&\textbf{NL}&\textbf{\textit{E}} \\
\cmidrule{2-35}
&&\multicolumn{11}{c}{\textbf{Sinhala - Tamil}}&\multicolumn{11}{c}{\textbf{English - Sinhala}}&\multicolumn{11}{c}{\textbf{English - Tamil}} \\
\toprule
\multicolumn{34}{l}{\textbf{CCMatrix}} \\
\hline
LASER3&BF&8\%&27\%&2\%&\textbf{37\%}&14\%&14\%&34\%&1\%&0\%&0\%&\textbf{63\%}&17\%&7\%&4\%&\textbf{28\%}&7\%&10\%&55\%&0\%&0\%&0\%&\textbf{72\%}&0\%&3\%&2\%&\textbf{5\%}&0\%&0\%&95\%&0\%&0\%&0\%&\textbf{95\%} \\
&AF&16\%&56\%&1\%&\textbf{73\%}&13\%&4\%&10\%&0\%&0\%&0\%&\textbf{27\%}&39\%&39\%&7\%&\textbf{85\%}&0\%&7\%&8\%&0\%&0\%&0\%&\textbf{15\%}&6\%&61\%&20\%&\textbf{87\%}&0\%&3\%&10\%&0\%&0\%&0\%&\textbf{13\%} \\
XLM-R&BF&1\%&10\%&0\%&\textbf{11\%}&40\%&19\%&29\%&0\%&1\%&0\%&\textbf{89\%}&1\%&0\%&0\%&\textbf{1\%}&13\%&4\%&80\%&2\%&0\%&0\%&\textbf{99\%}&0\%&0\%&2\%&\textbf{2\%}&3\%&5\%&90\%&0\%&0\%&0\%&\textbf{98\%} \\
&AF&0\%&20\%&2\%&\textbf{22\%}&13\%&29\%&35\%&0\%&1\%&0\%&\textbf{78\%}&3\%&8\%&26\%&\textbf{37\%}&0\%&2\%&53\%&8\%&0\%&0\%&\textbf{63\%}&0\%&39\%&31\%&\textbf{70\%}&1\%&3\%&21\%&4\%&0\%&1\%&\textbf{30\%} \\
LaBSE&BF&4\%&6\%&0\%&\textbf{10\%}&74\%&7\%&9\%&0\%&0\%&0\%&\textbf{90\%}&13\%&2\%&0\%&\textbf{15\%}&63\%&14\%&8\%&0\%&0\%&0\%&\textbf{85\%}&0\%&9\%&2\%&\textbf{11\%}&34\%&7\%&48\%&0\%&0\%&0\%&\textbf{89\%} \\
&AF&29\%&33\%&0\%&\textbf{62\%}&2\%&32\%&4\%&0\%&0\%&0\%&\textbf{38\%}&87\%&7\%&3\%&\textbf{97\%}&0\%&1\%&2\%&0\%&0\%&0\%&\textbf{3\%}&36\%&53\%&4\%&\textbf{93\%}&1\%&3\%&2\%&1\%&0\%&0\%&\textbf{7\%} \\
\hline
\multicolumn{34}{l}{\textbf{CCAligned}} \\
\hline
LASER3&BF&3\%&24\%&3\%&\textbf{30\%}&34\%&19\%&17\%&0\%&0\%&0\%&\textbf{70\%}&2\%&22\%&8\%&\textbf{32\%}&13\%&30\%&23\%&2\%&0\%&0\%&\textbf{68\%}&2\%&23\%&18\%&\textbf{43\%}&13\%&27\%&17\%&0\%&0\%&0\%&\textbf{57\%} \\
&AF&5\%&79\%&2\%&\textbf{86\%}&0\%&9\%&4\%&1\%&0\%&0\%&\textbf{14\%}&13\%&58\%&14\%&\textbf{85\%}&0\%&0\%&13\%&2\%&0\%&0\%&\textbf{15\%}&3\%&67\%&10\%&\textbf{80\%}&0\%&8\%&12\%&0\%&0\%&0\%&\textbf{20\%} \\
XLM-R&BF&0\%&0\%&2\%&\textbf{2\%}&48\%&49\%&0\%&0\%&0\%&1\%&\textbf{98\%}&2\%&0\%&0\%&\textbf{2\%}&72\%&20\%&6\%&0\%&0\%&0\%&\textbf{98\%}&0\%&8\%&4\%&\textbf{12\%}&42\%&16\%&15\%&8\%&0\%&7\%&\textbf{88\%} \\
&AF&20\%&33\%&4\%&\textbf{57\%}&1\%&22\%&19\%&0\%&1\%&0\%&\textbf{43\%}&18\%&18\%&20\%&\textbf{56\%}&0\%&6\%&34\%&4\%&0\%&0\%&\textbf{44\%}&6\%&46\%&30\%&\textbf{82\%}&0\%&9\%&9\%&0\%&0\%&0\%&\textbf{18\%} \\
LaBSE&BF&0\%&1\%&0\%&\textbf{1\%}&69\%&26\%&3\%&0\%&0\%&1\%&\textbf{99\%}&0\%&1\%&0\%&\textbf{1\%}&97\%&2\%&0\%&0\%&0\%&0\%&\textbf{99\%}&0\%&1\%&0\%&\textbf{1\%}&97\%&0\%&0\%&0\%&0\%&2\%&\textbf{99\%} \\
&AF&15\%&34\%&0\%&\textbf{49\%}&2\%&43\%&6\%&0\%&0\%&0\%&\textbf{51\%}&45\%&27\%&3\%&\textbf{75\%}&1\%&19\%&5\%&0\%&0\%&0\%&\textbf{25\%}&19\%&45\%&3\%&\textbf{67\%}&0\%&22\%&11\%&0\%&0\%&0\%&\textbf{33\%} \\
\bottomrule
\end{tabular}}
\vspace{1mm}
\caption{The Human evaluation results showing the average percentage for each annotation class for CCMatrix and CCAligned corpora for the En-Si, En-Ta and Si-Ta language pairs. The sample sentences have been obtained before (BF) and after (AF) applying the heuristics. \textbf{(\textit{C})} - overall correct percentage considering \textbf{CC} (perfect translation), \textbf{CN} (near perfect) and \textbf{CB} (boilerplate). \textbf{(\textit{E})} - overall error percentage considering \textbf{CCN} (Non-translatable overlaps), \textbf{CS} (correct but short sentence), \textbf{X} (wrong translation), \textbf{UN} (untranslated), \textbf{WL} (wrong language), \textbf{NL} (not a language).}\label{tab:HE_results}
\end{table*}

We hypothesise that some of these noisy sentences can be filtered using rule-based heuristics. Although applying heuristics is a common approach to improving the quality of parallel corpora~\cite{sloto-etal-2023-findings,steingrimsson-etal-2023-filtering}, the use of heuristics has not been consistent in the PDC tasks~\cite{steingrimsson-2023-sentence,bala2023improving,aulamo-etal-2020-opusfilter}. Previous research either applied a single heuristic or a subset of commonly used heuristics during pre-processing, with threshold choices varying across studies. Further, they did not systematically analyse the impact of heuristic-based filtering on the NMT performance.

In this research, we incorporate heuristics proposed in previous studies, along with a new heuristic of our own, and conduct an empirical study to investigate whether a more refined selection of top-ranked parallel sentences can be identified by systematically combining these heuristics.
\vspace{3mm}

\noindent Our key contributions are as follows:

\begin{itemize}
    \item We extend ~\citet{ranathunga-etal-2024-quality}'s parallel sentence categorization taxonomy with a new error category to capture an additional type of noise, which refers to sentence pairs with high-overlapping non-translatable text such as numbers, acronyms, URLs, etc.   
     \item {We empirically show that applying heuristics before ranking sentences based on embeddings derived from multiPLMs results in higher NMT scores, and reduces the disparity across multiPLMs.}
    \item We conduct a systematic study to analyse the impact of rule-based heuristics in filtering noisy sentences from the web-mined corpora and identify an optimal combination of heuristics that works across corpora and languages considered in the study.    
    \item We conduct a human evaluation to assess the impact of noise filtering across three multiPLMs.
\end{itemize}

\section{Related Work}\label{sec:related_work}

\subsection{MultiPLMs for PDC}

While employing a multiPLM for PDC is common, existing research experimented with only one multiPLM at a time. For example, in the WMT2023 shared task~\cite{sloto-etal-2023-findings}, LASER2 was utilised to set the task baseline, whereas for the same task,~\citet{steingrimsson-2023-sentence} used LaBSE.~\citet{gala2023indictrans2} also used LaBSE for their work. Therefore, the disparities across multiPLMs and biases specific to each multiPLM have not come to light.

On the other hand, studies conducted by~\citet{ranathunga-etal-2024-quality} and~\citet{moon2023doubts} reveal that using different multiPLMs for scoring and ranking parallel corpora, and training NMT models with the top-ranked corpora, results in a disparity.~\citet{moon2023doubts} observe that this is due to biases in multiPLMs, which tend to rank noisy parallel sentences highly. However, there has been no systematic study to identify these biases.

\subsection{Identifying Noise in Web-mined Corpora}

Recent research used categorical labels to annotate translation pairs, aiming at quantifying the noise types in web-mined corpora.~\citet{kreutzer-etal-2022-quality} used their taxonomy to conduct manual audits on random samples from three web-mined datasets and reported substantial noise, specifically for LRLs.~\citet{ranathunga-etal-2024-quality}'s taxonomy (See Table~\ref{tab:HE_error_taxonomy} in Appendix~\ref{sec:AppendixHumanEvaluation}) is an extension of~\citet{kreutzer-etal-2022-quality}'s taxonomy. They first ranked the datasets based on embeddings from a multiPLM, and then selected random samples from the top and bottom portions and conducted a quality audit. Their human evaluation reported that the quality of the parallel sentences varies heavily depending on the selected portion. These studies primarily focus on identifying general noise types; however, their effectiveness in quantifying the specific noise types to which multiPLMs are biased has not yet been evaluated to the best of our knowledge. 

\subsection{Heuristic-based PDC}\label{sec:heuristic_based_pdc}

In existing work, the commonly used rule-based heuristics can be categorised into four groups as described below: 

\noindent\textbf{Deduplication-based \textit{(Dedup)}: }Removing identical duplicates from the monolingual sides is a common practice~\cite{costa2022nllb}. Additionally deduplicating after removing non-alpha characters and punctuations~\cite{bala2023improving} could be found as its variants. While this step has been applied during the pre-processing stage, an empirical study has not been conducted to evaluate its impact on the final NMT performance. 

\noindent\textbf{Length-based (\textit{sLength}) : } \citet{gala2023indictrans2} and \citet{aulamo-etal-2023-unsupervised} have removed short sentences as a potential heuristic. Short sentences hinder NMT models in two ways~\cite{koehn-knowles-2017-six}: by providing insufficient syntactic and semantic information, and can result in an overfitting situation.    

\noindent\textbf{LID-based \textit{(LID)}:} Language  Identification is used to remove fully/partially untranslated text and content in the wrong language~\cite{steingrimsson-etal-2023-filtering,gala2023indictrans2,zhang-etal-2020-parallel-corpus}.

\noindent\textbf{Ratio-based :} Ratio-based heuristics identify and remove sentences that show significant structural imbalances between the source and target sentences. It is based on the assumption that, well-aligned sentence pairs tend to maintain consistent ratios in terms of character count, word count, or token distribution.
We observe three common types of ratio-based heuristics: (1)~source-to-target sentence length ratio (\textit{STRatio})~\citep{rossenbach-etal-2018-rwth,gale-church-1993-program}, (2)~alpha-only words to sentence words ratio (\textit{sentWRatio})~\citep{velayuthan-etal-2024-back,aulamo-etal-2020-opusfilter} and (3)~alpha-only character ratio with respect to the sentence characters (\textit{sentCRatio})~\cite{hangya-fraser-2018-unsupervised}. 

However, the impact of these heuristics in isolation and as a combination had not been evaluated systematically in the context of NMT.

\begin{table}[h]\centering
\scriptsize
\renewcommand{\arraystretch}{1.2}
\resizebox{1.0\linewidth}{!}{
\begin{tabular}{p{3cm}lp{6cm}}
\noalign{\hrule height 0.5pt}
\textbf{\citet{ranathunga-etal-2024-quality}} & \makecell{\textbf{Improved}\\
\textbf{Taxonomy}} & \textbf{Revision} \\
\noalign{\hrule height 0.5pt}
\multicolumn{1}{c}{-}&CCN &Perfect/near perfect translation where more than 30\% of the overlapping content is non-translatable such as numbers/acronyms/URLs/email etc.\\
\hline
Short Sentences (Max 3 words) &CS &Less than 5 words on either side \\
\hline
Wrong Language &WL & Specifically set a threshold as 30\% \\
\bottomrule
\end{tabular}}
\vspace{1mm}
\caption{A comparison of the improved taxonomy against~\citet{ranathunga-etal-2024-quality}'s (only showing the changes). See Table~\ref{tab:HE_error_taxonomy} in Appendix~\ref{sec:AppendixHumanEvaluation}  for the full taxonomy.}\label{tab:HE_error_taxonomy_comparison}
\end{table}

\section{Methodology}
\label{sec:methodology}
\textbf{Improved Taxonomy:} 
Although translation pairs can have overlapping URLs, acronyms, etc, excessive inclusions of such content in a sentence (e.g. consider the sentence \textit{`Contact:  Diane Anderson 076-8268914, info@sandnasbadenscamping.se'}\footnote{More examples are in Table~\ref{tab:ccn_example} in Appendix~\ref{sec:appendixImprovedNoiseTaxonomy}.}) do not provide meaningful content for an NMT system to learn from. However, under ~\citet{ranathunga-etal-2024-quality}'s taxonomy, such sentence pairs would likely be categorised as perfect translation-pairs \textbf{(CC)}. Therefore, we define a new noise category {\textbf{CCN} (high-overlapping non-translatable text) to capture such sentence pairs.

Secondly, we consider the upper limit for short sentences as five words\footnote{\label{note:thresholds}We considered thresholds 3,4 and 5 and empirically selected this threshold as it gave the highest NMT gains. Results in Table~\ref{tab:sLength_threshold} in Appendix~\ref{sec:AppendixSentLengthThresh}.}. Finally, we improve the definition of \textit{WL} (wrong language) to consider a threshold in determining whether a sentence pair should be marked as wrong language. Table~\ref{tab:HE_error_taxonomy_comparison} shows these changes. The complete list of these noise categories and example parallel sentences are available in Table~\ref{tab:HE_error_taxonomy} and Table~\ref{tab:examples_noisy_parallel_sentences} (respectively) in Appendix~\ref{sec:appendixImprovedNoiseTaxonomy}.

\paragraph{\textbf{Selection of Heuristics:}} Table~\ref{tab:heuristic2noise_mapping} shows how the heuristics discussed in Section~\ref{sec:heuristic_based_pdc} may help in removing different noise categories. Note that a deduplication-based heuristic cannot be associated with any in the taxonomy, as it does not apply to individual sentence pairs. In addition to the deduplication strategies discussed in Section~\ref{sec:heuristic_based_pdc}, we introduce an n-gram-based deduplication, meaning that sentences would be removed if they overlap in a consecutive n-gram text span.



\begin{table}[!htp]\centering
\scriptsize
\renewcommand{\arraystretch}{1.2}
\resizebox{1.0\linewidth}{!}{%
\begin{tabular}{L{4cm}ll}\toprule
\textbf{Noise Category}&\textbf{Short Label} &\textbf{Rule-based Heuristic} \\
\midrule
Not a language &NL &LID, sentWRatio, sentCRatio \\
Wrong language &WL &LID \\
Untranslated &UN &LID \\
Short Sentences &CS &sLength \\
High-overlapping non-translatable text &CCN &LID, sentWRatio, sentCRatio \\
Wrong translation &X &STRatio (With a length difference) \\
Boilerplate translation &CB &STRatio \\
\bottomrule
\end{tabular}}
\vspace{1mm}
\caption{Mapping between the noise category vs the noise mitigating heuristic.  }\label{tab:heuristic2noise_mapping}
\end{table}

\vspace{-10pt}
\paragraph{Human Evaluation} We conduct a human evaluation to quantify the noise before and after applying the heuristics. The annotator selection criteria, resources and training provided, the payment details, etc are described in  Appendix~\ref{sec:AppendixHumanEvaluation}.

For each language-pair, we obtain top 1000 samples in each of the ranked corpora using embeddings obtained from LASER3, XLM-R, and LaBSE, we randomly select 100 parallel sentences for each language pair. We ask the translators to annotate each sentence pair using the taxonomy discussed in Section~\ref{sec:methodology}. 

Each sentence pair is annotated by three translators to reduce any potential bias inherent in the individual translators. For the three annotators, the Fleiss Kappa scores are 0.833 for EnSi, 0.651 for EnTa, and 0.649 for  SiTa. We note that the results for EnTa and SiTa are very close.

\section{Experiments}\label{sec:experiments}
\subsection{Data} 
We use the language pairs, En-Si, En-Ta, and Si-Ta in our experiments. Sinhala and Tamil are morphologically rich, low-resource and mid-resourced languages~\cite{joshi-etal-2020-state, ranathunga-de-silva-2022-languages}, respectively. Languages are selected considering the availability of human evaluators. We select CCMatrix and CCAligned as the web-mined corpora. Both these corpora include parallel data for the language pairs considered in the research. 
More details of these datasets and language pairs are shown in Appendix~\ref{sec:appendixLangsDatasets}.

For the NMT experiments, we use the \textit{dev} and \textit{devtest} subsets from the Flores-200~\cite{costa2022nllb} dataset\footnote{https://github.com/openlanguagedata/flores} as validation and evaluation sets, respectively. Dataset statistics are provided in Table~\ref{tab:dataset_stats}.

\begin{table}[!htp]\centering
\scriptsize
\renewcommand{\arraystretch}{1.35}
\resizebox{0.49\textwidth}{!}{%
\begin{tabular}{lrrrrr}
\hline
\textbf{Language-pair} &\textbf{CCMatrix} &\textbf{CCAligned} &\textbf{dev} &\textbf{devtest} \\\midrule
En-Si &6,270,801 &619,711 &997 &1,012 \\
En-Ta &7,291,119 &880,547 &997 &1,012 \\
Si-Ta &215,966 &260,118 &997 &1,012 \\
\bottomrule
\end{tabular}}
\vspace{1mm}
\caption{Corpus statistics.}\label{tab:dataset_stats}
\end{table}

\begin{table*}[h]
\centering
\scriptsize
\renewcommand{\arraystretch}{1.1}
\resizebox{0.99\textwidth}{!}{
\setlength{\tabcolsep}{1pt}
\begin{tabular}{ll|cccccc|cccccc|cccccc}
\noalign{\hrule height 1.2pt}
\multirow{3}{*}{\textbf{Heuristic}} &\multirow{3}{*}{\shortstack{\textbf{Applicable}\\\textbf{Side}}}
&\multicolumn{6}{c}{\textbf{Sinhala-Tamil}} &\multicolumn{6}{c}{\textbf{English-Sinhala}} &\multicolumn{6}{c}{\textbf{English-Tamil}} \\
\cmidrule{3-20}
& &\multicolumn{3}{c}{\textbf{CCMatrix}} &\multicolumn{3}{c}{\textbf{CCAligned}} &\multicolumn{3}{c}{\textbf{CCMatrix}} &\multicolumn{3}{c}{\textbf{CCAligned}} &\multicolumn{3}{c}{\textbf{CCMatrix}} &\multicolumn{3}{c}{\textbf{CCAligned}} \\\cmidrule{3-20}
& &\textbf{LASER3} &\textbf{XLM-R} &\textbf{LaBSE} &\textbf{LASER3} &\textbf{XLM-R} &\textbf{LaBSE} &\textbf{LASER3} &\textbf{XLM-R} &\textbf{LaBSE} &\textbf{LASER3} &\textbf{XLM-R} &\textbf{LaBSE} &\textbf{LASER3} &\textbf{XLM-R} &\textbf{LaBSE} &\textbf{LASER3} &\textbf{XLM-R} &\textbf{LaBSE} \\
\midrule
Baseline & &31.08 &30.99 &31.63 &35.36 &35.97 &35.79 &30.76 &5.55 &14.49 &32.33 &19.39 &27.57 &19.02 &5.86 &14.20 &40.13 &17.40 &26.00 \\
\noalign{\hrule height 0.8pt}
DD &S &32.05 &31.50 &32.07 &36.40 &36.01 &34.98 &29.72 &6.35 &14.69 &33.26 &21.04 &28.22 &19.67 &4.93 &14.96 &40.87 &19.47 &26.26 \\
&T &31.39 &31.44 &31.73 &36.26 &35.86 &35.96 &33.81 &12.59 &25.97 &33.66 &21.41 &28.32 &19.48 &6.87 &17.96 &40.13 &17.90 &27.79 \\
&ST &32.26 &31.10 &32.25 &36.41 &36.08 &35.32 &34.01 &13.80 &26.18 &33.47 &22.22 &29.49 &20.32 &6.45 &17.53 &40.56 &19.83 &30.01 \\
DD-4gram &S &30.37 &30.65 &30.53 &35.74 &35.24 &34.55 &28.69 &8.56 &13.05 &31.56 &23.53 &28.25 &19.72 &7.06 &19.56 &35.54 &25.64 &26.49 \\
&T &31.00 &29.90 &29.39 &36.05 &35.98 &35.44 &31.79 &13.60 &23.66 &32.86 &24.95 &29.05 &19.82 &7.08 &20.23 &39.83 &27.44 &31.18 \\
&ST &30.86 &31.13 &30.80 &35.28 &35.36 &34.64 &28.72 &15.17 &20.45 &28.15 &15.45 &21.37 &18.15 &7.00 &21.37 &35.02 &25.70 &27.41 \\
DD-5gram &S &30.89 &30.90 &31.25 &35.64 &35.81 &35.87 &28.73 &7.14 &13.51 &33.44 &23.98 &28.79 &18.06 &4.70 &17.16 &40.39 &24.07 &29.07 \\
&T &31.24 &31.55 &32.10 &36.26 &35.87 &35.23 &33.98 &14.01 &26.23 &34.10 &22.27 &31.10 &20.15 &6.75 &18.78 &41.12 &24.05 &30.26 \\
&ST &30.78 &31.53 &31.35 &35.64 &35.94 &35.44 &31.95 &13.87 &23.07 &31.60 &17.10 &23.52 &19.61 &6.25 &20.12 &21.77 &25.22 &29.36 \\
DD-6gram &S &31.89 &30.82 &31.76 &36.31 &36.11 &35.88 &31.10 &7.62 &13.41 &33.53 &21.47 &28.51 &20.32 &5.47 &15.59 &40.48 &21.75 &27.64 \\
&T &32.51 &30.41 &32.29 &36.35 &36.23 &36.01 &34.21 &13.98 &24.91 &34.24 &23.63 &30.23 &\textbf{21.75} &6.69 &20.32 &40.44 &20.31 &30.48 \\
&ST &31.89 &30.82 &31.76 &35.84 &35.95 &35.54 &33.63 &14.96 &24.72 &33.29 &15.54 &25.55 &20.38 &7.18 &20.19 &41.73 &24.89 &31.06 \\
DD-7gram &S &31.48 &31.27 &32.03 &36.26 &35.67 &35.50 &30.93 &5.91 &15.94 &33.27 &19.90 &29.58 &21.54 &5.71 &16.49 &40.63 &20.01 &28.91 \\
&T &31.56 &31.06 &30.85 &36.44 &36.10 &35.16 &34.27 &13.72 &25.58 &32.97 &22.14 &28.22 &20.91 &7.37 &\textbf{21.96} &40.49 &19.18 &28.69 \\
&ST &31.48 &31.27 &32.03 &35.74 &35.90 &34.82 &33.93 &14.95 &24.95 &33.63 &14.58 &24.96 &17.56 &5.98 &20.71 &40.94 &22.16 &29.40 \\
DD+N &S &31.51 &31.37 &31.99 &36.61 &36.66 &35.99 &30.54 &5.92 &15.12 &34.77 &28.07 &31.81 &17.00 &5.60 &13.41 &41.40 &28.65 &35.22 \\
&T &31.17 &30.51 &32.09 &36.30 &36.45 &36.32 &33.83 &14.44 &25.86 &34.47 &27.27 &31.90 &17.54 &6.09 &19.01 &41.36 &28.40 &35.12 \\
&ST &31.71 &31.22 &31.66 &36.49 &36.37 &36.10 &33.83 &14.15 &26.12 &34.24 &28.45 &31.64 &19.19 &5.15 &18.92 &41.46 &30.49 &35.42 \\
DD+PN &S &31.90 &31.47 &31.02 &36.50 &36.00 &36.12 &30.55 &6.28 &16.67 &34.72 &27.25 &31.89 &18.15 &5.79 &15.66 &41.78 &30.55 &35.78 \\
&T &31.90 &32.05 &30.89 &36.63 &36.47 &\ul{\textbf{36.86}} &33.89 &14.81 &\textbf{26.31} &35.06 &27.69 &32.01 &21.57 &\textbf{8.24} &20.41 &41.64 &29.35 &35.32 \\
&ST &32.05 &31.31 &32.53 &35.96 &\textbf{36.71} &36.23 &33.37 &14.15 &26.08 &34.08 &27.80 &32.59 &20.99 &5.82 &18.83 &41.80 &30.69 &35.91 \\
DD+PN+4gram &ST+T &\multicolumn{3}{c}{\textbf{NA}} &\multicolumn{3}{c}{\textbf{NA}} &\multicolumn{3}{c}{\textbf{NA}} &30.64 &29.48 &30.19 &\multicolumn{3}{c}{\textbf{NA}} &41.82 &35.90 &\textbf{37.08} \\
DD+PN+5gram &ST + T &\colorbox{\cGreen}{\ul{\textbf{32.98}}} &\colorbox{\cGreen}{\ul{\textbf{32.73}}} &\colorbox{\cGreen}{\ul{\textbf{32.60}}} &36.24 &36.21 &36.35 &\ul{\textbf{34.50}} &\ul{\textbf{16.09}} &25.78 &33.81 &\textbf{30.33} &\textbf{32.74} &\multicolumn{3}{c}{\textbf{NA}} &\multicolumn{3}{c}{\textbf{NA}} \\
DD+PN+6gram &ST + T &30.41 &31.38 &31.42 &\ul{\textbf{36.73}} &36.62 &36.37 &\multicolumn{3}{c}{\textbf{NA}} &\textbf{35.24} &28.21 &31.26 &19.49 &6.67 &20.60 &\textbf{41.90} &\textbf{35.97} &35.94 \\
DD+PN+7gram &T +T &\multicolumn{3}{c}{\textbf{NA}} &\multicolumn{3}{c}{\textbf{NA}} &\multicolumn{3}{c}{\textbf{NA}} &\multicolumn{3}{c}{\textbf{NA}} &19.57 &7.55 &20.89 & & & \\
\noalign{\hrule height 0.8pt}
SL &\textbf{S} &\textbf{31.41} &\textbf{31.52} &\textbf{32.30} &36.42 &36.37 &36.52 &32.49 &6.58 &20.70 &33.86 &26.53 &32.97 &17.50 &5.11 &18.74 &41.40 &27.60 &36.77 \\
&T &31.38 &30.56 &31.97 &36.30 &\textbf{36.71} &36.58 &31.88 &7.83 &28.51 &\textbf{34.88} &29.42 &33.14 &18.52 &\textbf{6.33} &\textbf{21.73} &\textbf{41.54} &30.16 &37.61 \\
&ST &31.21 &31.32 &31.37 &\textbf{36.47} &35.99 &\textbf{36.60} &\textbf{32.82} &\textbf{8.24} &\ul{\textbf{29.96}} &34.83 &\textbf{29.55} &\ul{\textbf{33.50}} &\textbf{19.45} &5.33 &20.79 &41.14 &\textbf{32.67} &\ul{\textbf{38.08}} \\
\noalign{\hrule height 0.8pt}
LID &\textbf{S} &\textbf{31.48} &\textbf{31.36} &\textbf{31.78} &36.05 &36.03 &35.64 &31.00 &6.23 &14.69 &34.39 &27.33 &31.73 &18.44 &6.93 &13.43 &41.80 &31.41 &33.95 \\
&T &30.78 &31.14 &31.53 &35.68 &36.07 &35.85 &32.48 &12.22 &16.04 &33.70 &24.38 &30.48 &\textbf{29.59} &14.70 &24.24 &41.51 &24.24 &30.69 \\
&ST &31.43 &30.66 &31.40 &36.17 &36.12 &35.18 &31.99 &13.32 &\textbf{16.20} &34.11 &28.87 &32.26 &\textbf{29.59} &13.54 &23.45 &41.42 &32.33 &36.13 \\
LT &S &30.05 &31.25 &31.06 &35.60 &35.25 &34.29 &30.32 &7.12 &15.26 &\ul{\textbf{35.73}} &30.86 &32.69 &18.98 &6.02 &13.06 &41.60 &35.25 &36.29 \\
&T &31.28 &30.40 &30.68 &35.03 &30.01 &32.01 &32.82 &12.94 &15.81 &35.22 &27.46 &30.40 &\ul{\textbf{29.59}} &\ul{\textbf{15.24}} &24.51 &41.03 &30.01 &34.01 \\
&ST &30.33 &30.46 &30.71 &\ul{\textbf{36.73}} &\ul{\textbf{36.73}} &\textbf{36.80} &\textbf{32.84} &\textbf{14.08} &13.71 &35.11 &\ul{\textbf{32.97}} &\textbf{32.88} &28.93 &15.16 &\ul{\textbf{25.33}} &\ul{\textbf{42.63}} &\ul{\textbf{38.01}} &\textbf{37.40} \\
\noalign{\hrule height 0.8pt}
STRatio &- &31.74 &22.80 &31.34 &36.39 &35.74 &35.30 &31.09 &5.20 &15.40 &33.47 &24.05 &30.21 &\textbf{20.52} &5.40 &\textbf{18.29} &40.91 &22.71 &28.61 \\
sentWRatio &S &30.65 &30.62 &32.03 &36.17 &35.77 &35.54 &\textbf{31.50} &\textbf{7.40} &10.86 &\textbf{34.15} &25.97 &\textbf{31.35} &19.42 &5.79 &13.93 &\textbf{42.05} &29.70 &35.53 \\
&T &30.71 &31.59 &31.34 &36.24 &36.17 &\textbf{36.46} &30.99 &6.39 &15.13 &33.51 &26.93 &30.47 &18.61 &5.65 &11.08 &41.87 &30.06 &35.54 \\
&ST &31.93 &31.56 &30.98 &\textbf{36.44} &\textbf{36.72} &36.01 &30.64 &7.00 &\textbf{15.50} &33.85 &\textbf{28.73} &31.17 &18.99 &4.82 &14.08 &41.05 &\textbf{30.88} &\textbf{35.77} \\
sentCRatio &S &31.67 &31.24 &31.14 &35.94 &36.18 &35.86 &30.15 &7.05 &14.46 &34.06 &21.52 &30.10 &17.47 &6.22 &13.83 &40.68 &22.48 &29.37 \\
&T &30.98 &31.21 &31.93 &36.36 &35.43 &35.85 &30.65 &5.83 &15.28 &33.64 &23.14 &29.05 &19.90 &\textbf{6.78} &12.51 &40.78 &19.63 &29.42 \\
&\textbf{ST} &\textbf{32.28} &\textbf{31.90} &\textbf{32.04} &36.33 &35.60 &36.11 &30.85 &6.45 &14.64 &33.60 &23.84 &29.70 &19.54 &6.45 &10.79 &41.76 &21.82 &30.82 \\
\noalign{\hrule height 1.0pt}
\multicolumn{20}{l}{\textbf{Combined Heuristics}} \\
\noalign{\hrule height 1.0pt}
\multicolumn{20}{l}{DD+PN+ngram (SiTa-CCMatrix n=5, SiTa-CCAligned n= 7 EnSi-CCMatrix/CCAligned n=5, EnTa-CCMatrix n=7, EnTa-CCAligned n=6)}\\
+sLength &T + ST &30.17 &29.02 &29.99 &36.32 &36.81 &36.61 &35.03 &21.70 &26.32 &35.68 &33.49 &34.43 &30.29 &19.44 &29.85 &42.84 &39.36 &40.16 \\
+LT &T + ST &31.49 &30.13 &30.68 &36.58 &36.37 &37.02 &35.42 &19.58 &32.43 &34.77 &32.58 &34.72 &20.53 &7.52 &23.35 &42.68 &38.45 &39.60 \\
+sentWRatio &T+S &31.37 &30.55 &30.92 &\colorbox{\cGreen}{\textbf{36.83}} &36.75 &36.30 &33.99 &15.76 &24.92 &33.97 &31.40 &32.72 &21.67 &8.23 &24.58 &42.11 &37.47 &38.07 \\
+SL+LT &T + ST &29.28 &30.85 &29.96 &36.47 &36.81 &36.88 &35.70 &23.92 &32.77 &34.97 &34.92 &35.60 &30.65 &20.86 &31.49 &42.85 &41.17 &41.31 \\
+SL+sentWRatio &T + ST + ST &31.45 &\textbf{32.65} &31.17 &36.60 &\colorbox{\cGreen}{\textbf{36.85}} &36.32 &35.71 &18.93 &32.53 &35.45 &33.42 &33.82 &22.46 &9.11 &23.82 &41.97 &40.07 &40.06 \\
+SL+LT+sentWRatio &T+ST+ST+S &29.81 &29.53 &29.73 &\colorbox{\cGreen}{\textbf{36.83}} &36.66 &\colorbox{\cGreen}{\textbf{37.03}} &\colorbox{\cGreen}{\textbf{36.10}} &23.84 &\colorbox{\cGreen}{\textbf{33.94}} &36.15 &34.50 &\colorbox{\cGreen}{\textbf{35.67}} &\multicolumn{3}{c}{\textbf{NA}} &\colorbox{\cGreen}{\textbf{43.47}} &\colorbox{\cGreen}{\textbf{41.74}} &41.06 \\
+SL+LT+sentWRatio>0.8 &T+ST+ST+ST &28.70 &28.39 &28.34 &36.20 &36.60 &35.89 &35.66 &\colorbox{\cGreen}{\textbf{24.18}} &33.19 &\colorbox{\cGreen}{\textbf{36.26}} &\colorbox{\cGreen}{\textbf{35.66}} &35.42 &\multicolumn{3}{c}{\textbf{NA}} &42.08 &40.56 &\colorbox{\cGreen}{\textbf{42.02}} \\
+SL+LT+sentCRatio &T+ST+ST+ST &\textbf{32.64} &31.30 &\textbf{32.28} &\multicolumn{3}{c}{\textbf{NA}} &\multicolumn{3}{c}{\textbf{NA}} &\multicolumn{3}{c}{\textbf{NA}} &\multicolumn{3}{c}{\textbf{NA}} &\multicolumn{3}{c}{\textbf{NA}} \\
+SL+LT+STRatio &T+ST+ST+STR &\multicolumn{3}{c}{\textbf{NA}} &\multicolumn{3}{c}{\textbf{NA}} &\multicolumn{3}{c}{\textbf{NA}} &\multicolumn{3}{c}{\textbf{NA}} &\colorbox{\cGreen}{\textbf{30.67}} &\colorbox{\cGreen}{\textbf{23.36}} &\colorbox{\cGreen}{\textbf{31.80}}
&\multicolumn{3}{c}{\textbf{NA}} \\
\noalign{\hrule height 1.2pt}
\end{tabular}}
\vspace{1mm}
\caption{NMT results obtained after applying heuristics in isolation and in combination in the ablation study. The values in bold indicate the highest NMT score obtained for a given heuristic class or from the heuristic combination. The values underlined are the highest among the individual heuristics. Highlighted in green are the overall best values. Here \textbf{DD+PN} is \textit{deduplication+punctNums}, \textbf{SL} is \textit{sLength} and \textbf{LT} is \textit{LIDThresh}. \textbf{NA} would be when the particular experiment is not applicable for that language pair or the dataset.}\label{tab:results_full_table}
\end{table*}

\subsection{Selection of muliPLMs}
We select LASER3, XLM-R, and LaBSE for obtaining embeddings for the sentences to determine the semantic similarity. XLM-R, different to others, was trained purely on monolingual data, but has proven to be useful for cross-lingual tasks as well~\cite{choi2021analyzing,conneau-etal-2020-unsupervised}. All three models include En, Si, and Ta. Details of the multiPLMs are shown in Appendix~\ref{sec:AppendixCMultiPLMs}.

\subsection{Heuristic-based PDC Experiments}\label{exp_setup:dedup}
Each heuristic is applied independently to the source (S), target (T), and both sides (ST) of the corpora. In line with the original sentence alignment conducted for CCMatrix and CCAligned, we treat En as the source side. For Si-Ta, Si is considered the source because it is more common for a Si sentence to be translated to Ta~\cite{farhath2018integration}. Finally, for each multiPLM, the retained sentences are ranked in descending order with cosine similarity. 

\noindent\textbf{Deduplication}: We consider different granularities of deduplication. i.e. identical deduplication \textit{(dedup)}, deduplication after removing numbers only (\textit{nums}) and removing both numbers and punctuations (\textit{punctsNums}). Subsequently, we deduplicate considering different n-gram spans, i.e. 4-grams, 5-grams, 6-grams and 7-grams.

\noindent\textbf{Length-based:} We filter short sentences less than five words\footref{note:thresholds}. While some research has suggested removing extremely long sentences~\citep{minh-cong-etal-2023-fast,gala2023indictrans2}, we find that the percentage of longer sentences is lower and that removing them has a negligible effect. Thus, this result is not reported.

\noindent\textbf{LID-based:} We use a public LID model\footnote{https://github.com/facebookresearch/fairseq/tree/nllb}~\cite{costa2022nllb} to predict the language of each sentence. The predicted label is then used as a standalone heuristic (\textit{LID}) and in combination with its associated prediction probability~(\textit{LIDThresh}), with threshold of 0.7\footnote{Thresholds below 0.7 reduce NMT results.}.

\noindent\textbf{Ratio-based:} For~\textit{STRatio}, 0.79-1.39, 0.87-1.62 and 0.85-1.57 were selected as thresholds for En-Si, En-Ta and Si-Ta,  respectively. These thresholds are determined by calculating the mean and the standard deviation obtained for the validation set in a human-crafted trilingual dataset~\citep{fernando2020data,ranathunga2018si}. Following observations of~\citet{hangya-fraser-2018-unsupervised}, 0.6 is selected as the threshold for~\textit{sentWRatio} and ~\textit{sentCRatio}. 

\subsection{NMT Experiments}\label{sec:nmt_experiments}
First, a Sentencepiece\footnote{https://github.com/google/sentencepiece} tokenizer with a vocabulary size of 25000 is trained. Then we use the fairseq toolkit ~\citep{ott-etal-2019-fairseq} to model and train transformer-based Seq-to-Seq NMT models until convergence. Hyperparameters used in the NMT experiments are shown in Table~\ref{tab:nmt_hyperparameters} in Appendix~\ref{sec:AppendixNMTExperiments}. The baseline NMT models are trained on the top 100,000 sentence pairs from the ranked corpus. We use ChrF++~\cite{popovic-2017-chrf} to report NMT results.

\section{Results and Analysis}\label{sec:results}
We report the results obtained for the NMT models trained in the forward direction. The results of the experiments are in Table~\ref{tab:results_full_table}. Heuristic-wise best result is summarised in Table~\ref{tab:nmt_best_summary} in Appendix~\ref{sec:AppendixResultsAnaysis}.
As evident from these tables, as well as from Figure~\ref{fig:disparity}, NMT results across different multiPLMs show a great disparity in the baseline NMT scores for En-Si and En-Ta language pairs. In the following sub-sections, we discuss how the use of heuristics is useful in mitigating this disparity and improving overall NMT results.

\subsection{Impact of Heuristics on NMT Results}

\subsubsection{Impact of Deduplication-based PDC}

We observe that deduplication, irrespective of the heuristic-applied side (S/T/ST), outperforms the baseline in 89\% of the experiments. Overall, \textit{dedup} considering both source and target seems to be the most effective — it outperforms the baseline in 94\% of the experiments. In comparison, \textit{dedup} target-only and source-only outperform the baseline in 89\% and 83\% of the experiments, respectively. 

\begin{figure}[!htb]%
\centering
\includegraphics[trim={0cm 0cm 0cm 0cm}, clip, width=0.49\textwidth]{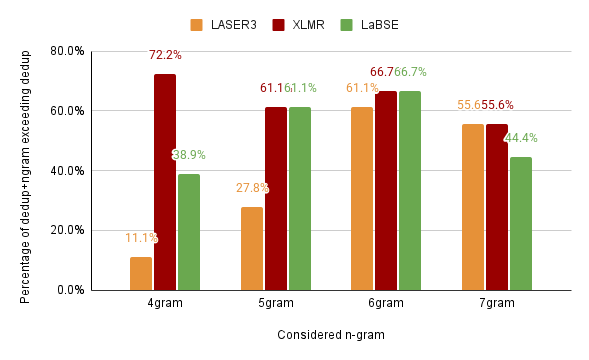}
\caption{Percentage of \textit{dedup+ngram} experiments exceeding the best result of \textit{dedup} for each~\textit{multiPLM}}\label{fig:ngram_vs_multiPLM}
\end{figure}

We apply our newly introduced n-gram-based deduplication~\textit{dedup+ngram}, on top of ~\textit{dedup}. We find that for each result column, there is a \textit{dedup+ngram} result that outperforms the best results obtained with the corresponding \textit{dedup} result. To observe the impact of the \textit{n} value on the NMT result, we plot  Figure~\ref{fig:ngram_vs_multiPLM} showing the percentage of  \textit{n-gram} experiments exceeding the highest \textit{dedup} result, with respect to the multiPLM (language-wise result is in Figure~\ref{fig:ngram_vs_langpair} of the Appendix~\ref{sec:AppendixResultsAnaysis}).
We observe a consistent pattern -  n=5 or 6 perform the best in a majority of cases. We believe that 4-gram results in an overly aggressive deduplication. However, the exact n-gram depends on the corpus characteristics.



We observe that~\textit{dedup+punctNums} outperforms ~\textit{dedup+nums} and~\textit{dedup} in 78\% and 89\% of the experiments 
(respectively), proving that~\textit{(dedup+punctNums)} to be more impactful. Finally, we analyse the impact of \textit{dedup+puntsNums+ngram}\footnote{ \textit{puntsNums} and \textit{ngram} has been applied on top of \textit{dedup}.}. Compared to other \textit{dedup} combinations\footnote{\textit{dedup}, \textit{dedup+ngram}, \textit{nums}, and \textit{dedup+punctNums}}, \textit{dedup+punctNums+ngram} produce the best result across 67\% of the experiments. 



\subsubsection{Impact of Length-Based 
 PDC}




\textit{sLength} surpasses the baseline in 89\% of the experiments. 
Therefore, we can conclude that \textit{sLength} is favourable as a heuristic. When analysing the side on which the heuristic is applied, we observe that applying it to both ST is the most effective (similar to \textit{dedup}), followed by T and then S (56\%, 28\%, and 17\% of the experiments, respectively). 

Recall that our manual inspection noticed that \textbf{XLM-R and LaBSE tend to prioritise shorter sentences}. This observation is affirmed by the \textit{sLength} results. For En-Si and En-Ta, this heuristic resulted in substantial gains for many of the XLM-R and LaBSE experiments, while it shows marginal improvements for LASER3.

\subsubsection{Impact of LID-Based 
 PDC} 
With LID-based PDC, we observe substantial improvements for XLM-R and LaBSE, except for Si-Ta, for which the gains are marginal. Gains are reported by LASER3 as well, though not very significant. The highest gain of +20.61 ChrF++ is reported for XLM-R for the CCAligned-EnTa corpus, while a gain of +11.40 is reported by LaBSE for the same corpus. 

NMT models trained after applying ~\textit{LID}  outperform the baseline in 85\% of the experiments, while ~\textit{LIDThresh} outperforms \textit{LID} in 72\% of the experiments. Therefore, we conclude that \textit{LIDThresh} would be the most suitable heuristic. We observe that the gains for \textit{LIDThresh} with respect to \textit{LID} are least with Si-Ta with 50\% while for En-Si and En-Ta it is 83\% each. We assume that this is due to a limitation with the LID model, which is not optimised for Si and Ta. 

\subsubsection{Impact of Ratio-based PDC}
We observe that~\textit{STRatio},~\textit{sentWRatio} and~\textit{sentCRatio} produce NMT gains over baseline for 56\%, 69\% and 80\% of experiments, respectively. 
Among the three ratio-based heuristics, \textit{sentWRatio} outperforms both \textit{STRatio} and \textit{sentCRatio} in 67\% of the experiments, making it the most effective for noise reduction. In contrast, \textit{STRatio} and \textit{sentCRatio} exceed the performance of the other two heuristics in only 11\% and 22\% of the cases, respectively.
Maximum gains are reported for the EnTa-CCAligned corpus, with +1.92, +13.48, and +9.77 ChrF++ for LASER3, XLM-R, and LaBSE respectively. 

\subsection{Summary of Heuristic-based PDC}
\paragraph{\textit{\textbf{1. Impact of the Individual Heuristics on the NMT Results:}}}The LID-based heuristic emerged as the most impactful in 44\% of the experiments. Deduplication and sentence-length heuristics were the most effective in 33\% and 17\% of the experiments, respectively. The ratio-based heuristic alone did not hold superior results compared to others, making it the least impactful single heuristic. There is no single heuristic that consistently produces the best gains for all scenarios. 
For LASER3, XLM-R, and LaBSE, we observe the highest ChrF++ gains of +10.57, +20.61, and +11.40 for CCMatrix-EnTa and CCAligned-EnTa and CCAligned-EnTa corpora, respectively.

\paragraph{\textit{\textbf{2. Impact of Combined Heuristics:}}}The combination of heuristics produced the best score, compared to the best-performing individual heuristics, except for the CCMatrix-SiTa. In this, the combination reduced the dataset size by 54\% (Table~\ref{tab:af_dataset_sizes} in Appendix~\ref{sec:AppendixResultsAnaysis}), resulting in only 98k parallel sentences. We suspect this reduction in dataset size and the residual noise could result in the reduced score.   

CCMatrix-SiTa gains across multiPLMs are marginal. The same pattern holds with LASER3 across En-Ta and En-Si languages, with the exception of CCMatrix-EnTa corpus. Since LASER3 is already trained with OPUS data~\cite{tiedemann-thottingal-2020-opus}\footnote{https://opus.nlpl.eu/}, it is believed that LASER3 is already optimised~\cite{moon2023doubts} towards ranking the CCMatrix and CCAligned corpora better. This is further affirmed with the relatively higher NMT scores of the baseline experiments.

For the rest of the ranked corpora, the best NMT scores were reported when the combined heuristic was applied. It was noted the combination always had heuristics, \textbf{\textit{dedup+punctNums+(n)gram+sLength+LIDThresh}} as a common combination for 77\% of the experiments. The specific \textbf{\textit{n}}-value was dependent on the dataset. In the combination producing the best gains, the ratio-based heuristic was different based on the dataset/language pair. In 61\% of the cases, the best scores were obtained with \textbf{\textit{sentWRatio}}, while \textbf{\textit{STRatio}} yielded the best results in 16\% of the experiments.

The combination performed best without the LID heuristic only in CCAligned-SiTa with XLM-R. However, we include \textit{LIDThresh} into the combination for two reasons. (1) Even with \textit{LIDThreshold}, the score only lags by (-0.19) ChrF++ scores compared to the best NMT score, which is negligible. (2) The most influential individual heuristic is LIDThresh-based for most cases. 

Therefore we recommend the heuristic combination, \textit{\textbf{dedup+punctNums+(n)gram+sLength}}\newline\textit{\textbf{+LIDThresh+sentWRatio}} to be applied as the rule-based heuristic combination for the PDC task.


\paragraph{\textit{\textbf{3. Reducing the Disparity Across multiPLMs: }}}We calculate the disparity ($\Delta$) as the difference in NMT scores between LASER3 and the XLM-R or LaBSE scores. Equation~\ref{eq:baseline_disparity} shows the baseline disparity calculation, where $BL_{\text{\textit{multiPLM}}}$ refers to the baseline NMT score obtained using embeddings from either XLM-R or LABSE during ranking. 

\vspace{-5mm}
\begin{equation}\label{eq:baseline_disparity}
\begin{minipage}{0.8\linewidth}
\centering
$\Delta_{\textit{Baseline}} = BL_{\text{\textit{LASER3}}} - BL_{\text{\textit{multiPLM}}}$
\end{minipage}
\end{equation}
\vspace{-5mm}

The disparity for each heuristic is calculated with respect to LASER3 and the best NMT score obtained from either XLM-R or LaBSE. The results are shown in Table~\ref{tab:nmt_disparity_analysis}.
Additionally, we calculate the disparity reduction percentage ($\Delta~Reduction(\%)$) as defined in Equation~\ref{eq:percentage_disparity}. Here, $\Delta_{heuristic}$ is the disparity corresponding to the best-performing NMT models after applying the respective heuristic or the optimal combination.

\begin{equation}\label{eq:percentage_disparity}
\begin{minipage}{0.9\linewidth}
$\Delta~Reduction(\%)
=\frac{\Delta_{\text{baseline}}-\Delta_{\text{heuristic}}}{\Delta_{\text{baseline}}}\times100\%$
\end{minipage}
\end{equation}

In Table~\ref{tab:nmt_disparity_analysis}\footnote{Since the disparity observed was marginal for the SiTa pair, we exclude this language pair from our discussion.}, we observe that the disparity has been reduced on average by 95.87\% for CCAligned-XLM-R/LaBSE and CCMatrix-LaBSE, compared to LASER3. The disparity between CCMatrix-XLM-R EnSi/EnTa models, compared to LASER3 was reduced by only 48\%, meaning that XLM-R ranked corpus contains noise that cannot be mitigated by the heuristics alone. However, our hypothesis holds in most cases, and it is safe to say that heuristic-based PDC mitigates the bias brought in by the multiPLM.

\begin{table}[!htp]\centering
\scriptsize
\renewcommand{\arraystretch}{1.2}
\resizebox{1.0\linewidth}{!}{%
\begin{tabular}{lrrrrr}\toprule
\multirow{2}{*}{\textbf{Heuristic}} &\multicolumn{2}{c}{\textbf{LASER3 vs XLM-R}} &\multicolumn{2}{c}{\textbf{LASER3 vs LaBSE}} \\\cmidrule{2-5}
&\shortstack[c]{\textbf{$\Delta$}\\ \textbf{(ChrF++)}} &
\shortstack[c]{\textbf{$\Delta$ Reduction}\\\textbf{(\%)}} &
\shortstack[c]{\textbf{$\Delta$}\\\textbf{(ChrF++)}}& \shortstack[c]{\textbf{$\Delta$ Reduction}\\\textbf{(\%)}} & \\
\midrule
\multicolumn{5}{c}{\textbf{CCMatrix}} \\
\hline
\multicolumn{5}{c}{\textbf{English - Sinhala}} \\
\hline
Baseline &25.21 & &16.27 & \\
Deduplication - based &18.41 &26.97\% &8.72 &46.40\% \\
Sentence Length - based &24.58 &2.50\% &2.86 &82.42\% \\
LID -based &18.76 &25.59\% &16.64 &-2.27\% \\
Ratio-based &24.10 &4.40\% &16.00 &1.66\% \\
Combined Heuristics &11.92 &\textbf{52.72\%} &2.16 &\textbf{86.72\%} \\
\hline
\multicolumn{5}{c}{\textbf{English - Tamil}} \\
\hline
Disparity &13.16 & &4.82 & \\
Reduction in disparity (dedup) &13.33 &-1.29\% &-0.39 &108.09\% \\
Reduction in disparity (sLength) &13.12 &0.30\% &-2.28 &147.30\% \\
LID &14.35 &-9.04\% &4.26 &11.62\% \\
Ratio-based &13.74 &-4.41\% &2.23 &53.73\% \\
Combined Heuristics &7.31 &\textbf{44.45\%} &-1.13 &\textbf{123.44\%} \\
\hline
\multicolumn{5}{c}{\textbf{CCAligned}} \\
\hline
\multicolumn{5}{c}{\textbf{English - Sinhala}} \\
\hline
Baseline &12.94 & &4.76 & \\
Deduplication Best &4.91 &62.06\% &2.50 &47.48\% \\
Reduction in disparity (sLength) &5.33 &58.81\% &1.38 &71.01\% \\
LID &2.76 &78.67\% &2.85 &40.13\% \\
Ratio-based &5.42 &58.11\% &2.80 &41.18\% \\
Combined Heuristics &0.60 &\textbf{95.36\%} &0.59 &\textbf{87.61\%} \\
\hline
\multicolumn{5}{c}{\textbf{English - Tamil}} \\
\hline
Disparity &22.73 & &14.13 & \\
Reduction in disparity (dedup) &5.93 &73.91\% &4.82 &65.89\% \\
Reduction in disparity (sLength) &8.87 &60.98\% &3.46 &75.51\% \\
LID &4.62 &79.67\% &5.23 &62.99\% \\
Ratio-based &11.17 &50.86\% &6.28 &55.56\% \\
Combined Heuristics &1.73 &\textbf{92.39\%} &1.45 &\textbf{89.74\%} \\
\bottomrule
\end{tabular}}\vspace{1mm}
\caption{\textbf{\textit{Disparity} ($\Delta$)} in ChrF++ points, among the NMT models (XLM-R/LaBSE) with the best scores with respect to LASER3 after applying the individual/combined heuristics. The \textbf{\textit{$\Delta$ Reduction (\%)}} is this disparity as a percentage of the baseline disparity.}\label{tab:nmt_disparity_analysis}
\end{table}



\subsection{Human Evaluation Results}\label{subsec:human_eval_results}
As shown in Table~\ref{tab:HE_results}, heuristic-based PDC had reduced the noise in the top-ranked samples consistently, irrespective of the language pair and the considered multiPLM. Some gains are quite significant — for example, the amount of correct pairs~(C) improvement for  CCMatrix-EnSi (LaBSE), CCMatrix-EnTa (LaBSE), and CCAligned-EnSi (LaBSE) were 82\%, 82\%, and 74\% respectively. 
%
CS, CCN and UN noise categories are noted to be contributing towards this noise percentage. However, after heuristic filtration, the error drops drastically. On average, for LASER, XLM-R, and LaBSE, the final error percentages are 2.17\%, 2.50\% and 1.0\%, respectively. The evaluation further reveals that the residual noise are from untranslated (UN) and overlapping text (CCN) classes. These findings reveal that rule-based heuristics, are not effective in eliminating those types of noise. As a result, we would need to employ an alignment model similar to~\citet{steingrimsson-etal-2023-filtering} or~\citet{minh-cong-etal-2023-fast} to remove such residual noise from the corpus.

In conclusion, the human evaluation results indicate that the heuristic-based PDC approach is beneficial for parallel sentence ranking in two key ways.
First, it produces the top-ranked sentence pairs from multiPLM to be qualitatively comparable. Secondly, it removes the noisy parallel sentences that cause the disparity among the NMT systems trained using ranked corpora based on multiPLMs.

\section{NMT Performance based on Training Data Size}
We further investigate the effect of the training dataset size on the final NMT performance. We sampled the top 50k, 100k, 150k, and 200k from each of the ranked corpora after applying the heuristic-combination, and train NMT models. Results are in Figure~\ref{fig:size_vs_nmt} of the Appendix~\ref{sec:AppendixNMTVsTrainingData}.

For Si-Ta, moderate but consistent improvements are observed from 50K to 100K samples for both CCMatrix and CCAligned corpora, irrespective of the multiPLMs. However, after heuristic-based filtering, the resulting Si-Ta parallel corpus contained only about 100K sentence pairs (Table~\ref{tab:af_dataset_sizes} in Appendix~\ref{sec:AppendixResultsAnaysis}). Therefore, NMT experiments with 150K and 200K sentence pairs could not be conducted due to insufficient data.

For En-Ta,  NMT scores declined after 100K, while for En-Si, an improvement was observed until 200K, except for the CCMatrix ranked using XLM-R embeddings. Hence, we suspect that noise may be more prominent beyond 100K for En-Ta.  
In conclusion, increasing corpus size can enhance NMT performance, but only when the underlying data is sufficiently clean and well-aligned.
\section{Conclusion}\label{sec:conclusion}

In this research, we empirically analysed the disparity between the NMT systems trained with the web-mined corpora ranked using embeddings derived from multiPLMs.  With a human evaluation, we showed that this disparity is due to different types of noise creeping into the top-ranked portion of corpora when different multiPLMs are used. 

We made use of rule-based heuristics to remove this noise. After a systematic evaluation of heuristics, we were able to identify optimal heuristic combinations that resulted in higher NMT scores. Therefore, for anyone planning to use web-mined corpora, our recommendation is to first filter out noisy sentences using heuristics and then to do ranking on the embeddings derived from the multiPLM. In contrast to the recent PDC work~\cite{steingrimsson-2023-sentence,minh-cong-etal-2023-fast} that employs several deep learning model-based rigorous filtration pipelines, our technique is much simpler. 

Human evaluation indicated that even after applying heuristics, some noise remains. In future, we plan to apply techniques such as classification-based approaches to remove such noise, or to apply translation post-editing to improve the sentence quality.  



\section{Limitations and Ethical Concerns}\label{sec:limitations}

\subsection{Limitations}
We found that the LID was suboptimal for identifying Si or Ta languages. Therefore, such models would need to be optimised or better LID models would need to be used, such that they are effective in predicting the language label correctly.
Furthermore, due to the lack of human annotators, we had to limit this analysis to only three language pairs. We can extend this study to more low-resource language pairs provided that we can find suitable annotators.

\subsection{Ethical Concerns}

We use publicly available datasets. Due to the dataset size, we did not have the resources or funding to check all the curated parallel sentences manually for offensive content.~\citet{fernando2020data} provided the human-crafted dataset to be used in this research. Further, we do not disclose or share any personal details of the annotators publicly or with any other party. We have offered the annotators the standard rate and have settled the payments in full. The only details we disclose are in Appendix~\ref{sec:AppendixHumanEvaluation}.

\section*{Acknowledgments}
This research was funded by the Google Award for Inclusion Research (AIR) 2022 received by Surangika Ranathunga and Nisansa de Silva. We would also like to thank and acknowledge the National Languages Processing Centre (NLPC), at the  University of Moratuwa for providing the GPUs to execute the experiments related to the research. 

\bibliography{anthology,references}

\begin{thebibliography}{43}
\providecommand{\natexlab}[1]{#1}

\bibitem[{Aulamo et~al.(2023)Aulamo, de~Gibert, Virpioja, and Tiedemann}]{aulamo-etal-2023-unsupervised}
Mikko Aulamo, Ona de~Gibert, Sami Virpioja, and J{\"o}rg Tiedemann. 2023.
\newblock \href {https://aclanthology.org/2023.eamt-1.4/} {Unsupervised feature selection for effective parallel corpus filtering}.
\newblock In \emph{Proceedings of the 24th Annual Conference of the European Association for Machine Translation}, pages 31--38. European Association for Machine Translation.

\bibitem[{Aulamo et~al.(2020)Aulamo, Virpioja, and Tiedemann}]{aulamo-etal-2020-opusfilter}
Mikko Aulamo, Sami Virpioja, and J{\"o}rg Tiedemann. 2020.
\newblock \href {https://doi.org/10.18653/v1/2020.acl-demos.20} {{O}pus{F}ilter: A configurable parallel corpus filtering toolbox}.
\newblock In \emph{Proceedings of the 58th Annual Meeting of the Association for Computational Linguistics: System Demonstrations}, pages 150--156. Association for Computational Linguistics.

\bibitem[{Bala~Das et~al.(2023)Bala~Das, Biradar, Kumar~Mishra, and Kr.~Patra}]{bala2023improving}
Sudhansu Bala~Das, Atharv Biradar, Tapas Kumar~Mishra, and Bidyut Kr.~Patra. 2023.
\newblock Improving multilingual neural machine translation system for indic languages.
\newblock \emph{ACM Transactions on Asian and Low-Resource Language Information Processing}, 22(6):1--24.

\bibitem[{Bane et~al.(2022)Bane, Uguet, Stribi{\.z}ew, and Zaretskaya}]{bane-etal-2022-comparison}
Fred Bane, Celia~Soler Uguet, Wiktor Stribi{\.z}ew, and Anna Zaretskaya. 2022.
\newblock \href {https://aclanthology.org/2022.amta-upg.22/} {A comparison of data filtering methods for neural machine translation}.
\newblock In \emph{Proceedings of the 15th Biennial Conference of the Association for Machine Translation in the Americas (Volume 2: Users and Providers Track and Government Track)}, pages 313--325. Association for Machine Translation in the Americas.

\bibitem[{Ba{\~n}{\'o}n et~al.(2020)Ba{\~n}{\'o}n, Chen, Haddow, Heafield, Hoang, Espl{\`a}-Gomis, Forcada, Kamran, Kirefu, Koehn, Ortiz~Rojas, Pla~Sempere, Ram{\'i}rez-S{\'a}nchez, Sarr{\'i}as, Strelec, Thompson, Waites, Wiggins, and Zaragoza}]{banon-etal-2020-paracrawl2}
Marta Ba{\~n}{\'o}n, Pinzhen Chen, Barry Haddow, Kenneth Heafield, Hieu Hoang, Miquel Espl{\`a}-Gomis, Mikel~L. Forcada, Amir Kamran, Faheem Kirefu, Philipp Koehn, Sergio Ortiz~Rojas, Leopoldo Pla~Sempere, Gema Ram{\'i}rez-S{\'a}nchez, Elsa Sarr{\'i}as, Marek Strelec, Brian Thompson, William Waites, Dion Wiggins, and Jaume Zaragoza. 2020.
\newblock \href {https://doi.org/10.18653/v1/2020.acl-main.417} {{P}ara{C}rawl: Web-scale acquisition of parallel corpora}.
\newblock In \emph{Proceedings of the 58th Annual Meeting of the Association for Computational Linguistics}, pages 4555--4567, Online. Association for Computational Linguistics.

\bibitem[{Chaudhary et~al.(2019)Chaudhary, Tang, Guzm{\'a}n, Schwenk, and Koehn}]{chaudhary-etal-2019-low}
Vishrav Chaudhary, Yuqing Tang, Francisco Guzm{\'a}n, Holger Schwenk, and Philipp Koehn. 2019.
\newblock \href {https://doi.org/10.18653/v1/W19-5435} {Low-resource corpus filtering using multilingual sentence embeddings}.
\newblock In \emph{Proceedings of the Fourth Conference on Machine Translation (Volume 3: Shared Task Papers, Day 2)}, pages 261--266. Association for Computational Linguistics.

\bibitem[{Choi et~al.(2021)Choi, Kim, Joe, Min, and Gwon}]{choi2021analyzing}
Hyunjin Choi, Judong Kim, Seongho Joe, Seungjai Min, and Youngjune Gwon. 2021.
\newblock Analyzing zero-shot cross-lingual transfer in supervised nlp tasks.
\newblock In \emph{2020 25th International Conference on Pattern Recognition (ICPR)}, pages 9608--9613. IEEE.

\bibitem[{Conneau et~al.(2020)Conneau, Khandelwal, Goyal, Chaudhary, Wenzek, Guzm{\'a}n, Grave, Ott, Zettlemoyer, and Stoyanov}]{conneau-etal-2020-unsupervised}
Alexis Conneau, Kartikay Khandelwal, Naman Goyal, Vishrav Chaudhary, Guillaume Wenzek, Francisco Guzm{\'a}n, Edouard Grave, Myle Ott, Luke Zettlemoyer, and Veselin Stoyanov. 2020.
\newblock \href {https://doi.org/10.18653/v1/2020.acl-main.747} {Unsupervised cross-lingual representation learning at scale}.
\newblock In \emph{Proceedings of the 58th Annual Meeting of the Association for Computational Linguistics}, pages 8440--8451. Association for Computational Linguistics.

\bibitem[{Costa-juss{\`a} et~al.(2022)Costa-juss{\`a}, Cross, {\c{C}}elebi, Elbayad, Heafield, Heffernan, Kalbassi, Lam, Licht, Maillard et~al.}]{costa2022nllb}
Marta~R Costa-juss{\`a}, James Cross, Onur {\c{C}}elebi, Maha Elbayad, Kenneth Heafield, Kevin Heffernan, Elahe Kalbassi, Janice Lam, Daniel Licht, Jean Maillard, et~al. 2022.
\newblock No language left behind: Scaling human-centered machine translation.
\newblock \emph{arXiv preprint arXiv:2207.04672}.

\bibitem[{de~Silva(2025)}]{de2025survey}
Nisansa de~Silva. 2025.
\newblock {Survey on Publicly Available Sinhala Natural Language Processing Tools and Research}.
\newblock \emph{arXiv preprint arXiv:1906.02358v25}.

\bibitem[{El-Kishky et~al.(2020)El-Kishky, Chaudhary, Guzm{\'a}n, and Koehn}]{el-kishky-etal-2020-ccaligned}
Ahmed El-Kishky, Vishrav Chaudhary, Francisco Guzm{\'a}n, and Philipp Koehn. 2020.
\newblock \href {https://doi.org/10.18653/v1/2020.emnlp-main.480} {{CCA}ligned: A massive collection of cross-lingual web-document pairs}.
\newblock In \emph{Proceedings of the 2020 Conference on Empirical Methods in Natural Language Processing (EMNLP)}, pages 5960--5969. Association for Computational Linguistics.

\bibitem[{Farhath et~al.(2018)Farhath, Ranathunga, Jayasena, and Dias}]{farhath2018integration}
Fathima Farhath, Surangika Ranathunga, Sanath Jayasena, and Gihan Dias. 2018.
\newblock Integration of bilingual lists for domain-specific statistical machine translation for sinhala-tamil.
\newblock In \emph{2018 Moratuwa Engineering Research Conference (MERCon)}, pages 538--543. IEEE.

\bibitem[{Feng et~al.(2022)Feng, Yang, Cer, Arivazhagan, and Wang}]{feng-etal-2022-language}
Fangxiaoyu Feng, Yinfei Yang, Daniel Cer, Naveen Arivazhagan, and Wei Wang. 2022.
\newblock \href {https://doi.org/10.18653/v1/2022.acl-long.62} {Language-agnostic {BERT} sentence embedding}.
\newblock In \emph{Proceedings of the 60th Annual Meeting of the Association for Computational Linguistics (Volume 1: Long Papers)}, pages 878--891. Association for Computational Linguistics.

\bibitem[{Fernando et~al.(2020)Fernando, Ranathunga, and Dias}]{fernando2020data}
Aloka Fernando, Surangika Ranathunga, and Gihan Dias. 2020.
\newblock Data augmentation and terminology integration for domain-specific sinhala-english-tamil statistical machine translation.
\newblock \emph{arXiv preprint arXiv:2011.02821}.

\bibitem[{Gala et~al.(2023)Gala, Chitale, AK, Gumma, Doddapaneni, Kumar, Nawale, Sujatha, Puduppully, Raghavan et~al.}]{gala2023indictrans2}
Jay Gala, Pranjal~A Chitale, Raghavan AK, Varun Gumma, Sumanth Doddapaneni, Aswanth Kumar, Janki Nawale, Anupama Sujatha, Ratish Puduppully, Vivek Raghavan, et~al. 2023.
\newblock Indictrans2: Towards high-quality and accessible machine translation models for all 22 scheduled indian languages.
\newblock \emph{arXiv preprint arXiv:2305.16307}.

\bibitem[{Gale and Church(1993)}]{gale-church-1993-program}
William~A. Gale and Kenneth~W. Church. 1993.
\newblock \href {https://aclanthology.org/J93-1004/} {A program for aligning sentences in bilingual corpora}.
\newblock \emph{Computational Linguistics}, 19(1):75--102.

\bibitem[{Hangya and Fraser(2018)}]{hangya-fraser-2018-unsupervised}
Viktor Hangya and Alexander Fraser. 2018.
\newblock \href {https://doi.org/10.18653/v1/W18-6477} {An unsupervised system for parallel corpus filtering}.
\newblock In \emph{Proceedings of the Third Conference on Machine Translation: Shared Task Papers}, pages 882--887. Association for Computational Linguistics.

\bibitem[{Heffernan et~al.(2022)Heffernan, {\c{C}}elebi, and Schwenk}]{heffernan-etal-2022-bitext}
Kevin Heffernan, Onur {\c{C}}elebi, and Holger Schwenk. 2022.
\newblock \href {https://doi.org/10.18653/v1/2022.findings-emnlp.154} {Bitext mining using distilled sentence representations for low-resource languages}.
\newblock In \emph{Findings of the Association for Computational Linguistics: EMNLP 2022}, pages 2101--2112. Association for Computational Linguistics.

\bibitem[{Herold et~al.(2022)Herold, Rosendahl, Vanvinckenroye, and Ney}]{herold-etal-2022-detecting}
Christian Herold, Jan Rosendahl, Joris Vanvinckenroye, and Hermann Ney. 2022.
\newblock \href {https://doi.org/10.18653/v1/2022.findings-acl.200} {Detecting various types of noise for neural machine translation}.
\newblock In \emph{Findings of the Association for Computational Linguistics: ACL 2022}, pages 2542--2551. Association for Computational Linguistics.

\bibitem[{Joshi et~al.(2020)Joshi, Santy, Budhiraja, Bali, and Choudhury}]{joshi-etal-2020-state}
Pratik Joshi, Sebastin Santy, Amar Budhiraja, Kalika Bali, and Monojit Choudhury. 2020.
\newblock \href {https://doi.org/10.18653/v1/2020.acl-main.560} {The state and fate of linguistic diversity and inclusion in the {NLP} world}.
\newblock In \emph{Proceedings of the 58th Annual Meeting of the Association for Computational Linguistics}, pages 6282--6293. Association for Computational Linguistics.

\bibitem[{Khayrallah and Koehn(2018)}]{khayrallah-koehn-2018-impact}
Huda Khayrallah and Philipp Koehn. 2018.
\newblock \href {https://doi.org/10.18653/v1/W18-2709} {On the impact of various types of noise on neural machine translation}.
\newblock In \emph{Proceedings of the 2nd Workshop on Neural Machine Translation and Generation}, pages 74--83. Association for Computational Linguistics.

\bibitem[{Koehn et~al.(2020)Koehn, Chaudhary, El-Kishky, Goyal, Chen, and Guzm{\'a}n}]{koehn-etal-2020-findings}
Philipp Koehn, Vishrav Chaudhary, Ahmed El-Kishky, Naman Goyal, Peng-Jen Chen, and Francisco Guzm{\'a}n. 2020.
\newblock \href {https://aclanthology.org/2020.wmt-1.78/} {Findings of the {WMT} 2020 shared task on parallel corpus filtering and alignment}.
\newblock In \emph{Proceedings of the Fifth Conference on Machine Translation}, pages 726--742. Association for Computational Linguistics.

\bibitem[{Koehn et~al.(2019)Koehn, Guzm{\'a}n, Chaudhary, and Pino}]{koehn-etal-2019-findings}
Philipp Koehn, Francisco Guzm{\'a}n, Vishrav Chaudhary, and Juan Pino. 2019.
\newblock \href {https://doi.org/10.18653/v1/W19-5404} {Findings of the {WMT} 2019 shared task on parallel corpus filtering for low-resource conditions}.
\newblock In \emph{Proceedings of the Fourth Conference on Machine Translation (Volume 3: Shared Task Papers, Day 2)}, pages 54--72. Association for Computational Linguistics.

\bibitem[{Koehn and Knowles(2017)}]{koehn-knowles-2017-six}
Philipp Koehn and Rebecca Knowles. 2017.
\newblock \href {https://doi.org/10.18653/v1/W17-3204} {Six challenges for neural machine translation}.
\newblock In \emph{Proceedings of the First Workshop on Neural Machine Translation}, pages 28--39. Association for Computational Linguistics.

\bibitem[{Kreutzer et~al.(2022)Kreutzer, Caswell, Wang, Wahab, van Esch, Ulzii-Orshikh, Tapo, Subramani, Sokolov, Sikasote, Setyawan, Sarin, Samb, Sagot, Rivera, Rios, Papadimitriou, Osei, Suarez, Orife, Ogueji, Rubungo, Nguyen, M{\"u}ller, M{\"u}ller, Muhammad, Muhammad, Mnyakeni, Mirzakhalov, Matangira, Leong, Lawson, Kudugunta, Jernite, Jenny, Firat, Dossou, Dlamini, de~Silva, {\c{C}}abuk~Ball{\i}, Biderman, Battisti, Baruwa, Bapna, Baljekar, Azime, Awokoya, Ataman, Ahia, Ahia, Agrawal, and Adeyemi}]{kreutzer-etal-2022-quality}
Julia Kreutzer, Isaac Caswell, Lisa Wang, Ahsan Wahab, Daan van Esch, Nasanbayar Ulzii-Orshikh, Allahsera Tapo, Nishant Subramani, Artem Sokolov, Claytone Sikasote, Monang Setyawan, Supheakmungkol Sarin, Sokhar Samb, Beno{\^i}t Sagot, Clara Rivera, Annette Rios, Isabel Papadimitriou, Salomey Osei, Pedro~Ortiz Suarez, Iroro Orife, Kelechi Ogueji, Andre~Niyongabo Rubungo, Toan~Q. Nguyen, Mathias M{\"u}ller, Andr{\'e} M{\"u}ller, Shamsuddeen~Hassan Muhammad, Nanda Muhammad, Ayanda Mnyakeni, Jamshidbek Mirzakhalov, Tapiwanashe Matangira, Colin Leong, Nze Lawson, Sneha Kudugunta, Yacine Jernite, Mathias Jenny, Orhan Firat, Bonaventure F.~P. Dossou, Sakhile Dlamini, Nisansa de~Silva, Sakine {\c{C}}abuk~Ball{\i}, Stella Biderman, Alessia Battisti, Ahmed Baruwa, Ankur Bapna, Pallavi Baljekar, Israel~Abebe Azime, Ayodele Awokoya, Duygu Ataman, Orevaoghene Ahia, Oghenefego Ahia, Sweta Agrawal, and Mofetoluwa Adeyemi. 2022.
\newblock \href {https://doi.org/10.1162/tacl\_a\_00447} {Quality at a glance: An audit of web-crawled multilingual datasets}.
\newblock \emph{Transactions of the Association for Computational Linguistics}, 10:50--72.

\bibitem[{Minh-Cong et~al.(2023)Minh-Cong, Vinh, and Le-Minh}]{minh-cong-etal-2023-fast}
Nguyen-Hoang Minh-Cong, Nguyen~Van Vinh, and Nguyen Le-Minh. 2023.
\newblock \href {https://doi.org/10.18653/v1/2023.wmt-1.37} {A fast method to filter noisy parallel data {WMT}2023 shared task on parallel data curation}.
\newblock In \emph{Proceedings of the Eighth Conference on Machine Translation}, pages 359--365. Association for Computational Linguistics.

\bibitem[{Moon et~al.(2023)Moon, Park, Koo, Lee, Lee, Seo, Eo, Jang, Kim, Lee et~al.}]{moon2023doubts}
Hyeonseok Moon, Chanjun Park, Seonmin Koo, Jungseob Lee, Seungjun Lee, Jaehyung Seo, Sugyeong Eo, Yoonna Jang, Hyunjoong Kim, Hyoung-gyu Lee, et~al. 2023.
\newblock Doubts on the reliability of parallel corpus filtering.
\newblock \emph{Expert Systems with Applications}, 233:120962.

\bibitem[{Ott et~al.(2019)Ott, Edunov, Baevski, Fan, Gross, Ng, Grangier, and Auli}]{ott-etal-2019-fairseq}
Myle Ott, Sergey Edunov, Alexei Baevski, Angela Fan, Sam Gross, Nathan Ng, David Grangier, and Michael Auli. 2019.
\newblock \href {https://doi.org/10.18653/v1/N19-4009} {fairseq: A fast, extensible toolkit for sequence modeling}.
\newblock In \emph{Proceedings of the 2019 Conference of the North {A}merican Chapter of the Association for Computational Linguistics (Demonstrations)}, pages 48--53. Association for Computational Linguistics.

\bibitem[{Papineni et~al.(2002)Papineni, Roukos, Ward, and Zhu}]{papineni-etal-2002-bleu}
Kishore Papineni, Salim Roukos, Todd Ward, and Wei-Jing Zhu. 2002.
\newblock \href {https://doi.org/10.3115/1073083.1073135} {{B}leu: a method for automatic evaluation of machine translation}.
\newblock In \emph{Proceedings of the 40th Annual Meeting of the Association for Computational Linguistics}, pages 311--318. Association for Computational Linguistics.

\bibitem[{Popovi{\'c}(2017)}]{popovic-2017-chrf}
Maja Popovi{\'c}. 2017.
\newblock \href {https://doi.org/10.18653/v1/W17-4770} {chr{F}++: words helping character n-grams}.
\newblock In \emph{Proceedings of the Second Conference on Machine Translation}, pages 612--618. Association for Computational Linguistics.

\bibitem[{Post(2018)}]{post-2018-call}
Matt Post. 2018.
\newblock \href {https://doi.org/10.18653/v1/W18-6319} {A call for clarity in reporting {BLEU} scores}.
\newblock In \emph{Proceedings of the Third Conference on Machine Translation: Research Papers}, pages 186--191. Association for Computational Linguistics.

\bibitem[{Ranathunga and de~Silva(2022)}]{ranathunga-de-silva-2022-languages}
Surangika Ranathunga and Nisansa de~Silva. 2022.
\newblock \href {https://doi.org/10.18653/v1/2022.aacl-main.62} {Some languages are more equal than others: Probing deeper into the linguistic disparity in the {NLP} world}.
\newblock In \emph{Proceedings of the 2nd Conference of the Asia-Pacific Chapter of the Association for Computational Linguistics and the 12th International Joint Conference on Natural Language Processing (Volume 1: Long Papers)}, pages 823--848. Association for Computational Linguistics.

\bibitem[{Ranathunga et~al.(2024)Ranathunga, De~Silva, Menan, Fernando, and Rathnayake}]{ranathunga-etal-2024-quality}
Surangika Ranathunga, Nisansa De~Silva, Velayuthan Menan, Aloka Fernando, and Charitha Rathnayake. 2024.
\newblock \href {https://doi.org/10.18653/v1/2024.eacl-long.52} {Quality does matter: A detailed look at the quality and utility of web-mined parallel corpora}.
\newblock In \emph{Proceedings of the 18th Conference of the European Chapter of the Association for Computational Linguistics (Volume 1: Long Papers)}, pages 860--880. Association for Computational Linguistics.

\bibitem[{Ranathunga et~al.(2018)Ranathunga, Farhath, Thayasivam, Jayasena, and Dias}]{ranathunga2018si}
Surangika Ranathunga, Fathima Farhath, Uthayasanker Thayasivam, Sanath Jayasena, and Gihan Dias. 2018.
\newblock Si-ta: Machine translation of sinhala and tamil official documents.
\newblock In \emph{2018 National Information Technology Conference (NITC)}, pages 1--6. IEEE.

\bibitem[{Rossenbach et~al.(2018)Rossenbach, Rosendahl, Kim, Gra{\c{c}}a, Gokrani, and Ney}]{rossenbach-etal-2018-rwth}
Nick Rossenbach, Jan Rosendahl, Yunsu Kim, Miguel Gra{\c{c}}a, Aman Gokrani, and Hermann Ney. 2018.
\newblock \href {https://doi.org/10.18653/v1/W18-6487} {The {RWTH} {A}achen {U}niversity filtering system for the {WMT} 2018 parallel corpus filtering task}.
\newblock In \emph{Proceedings of the Third Conference on Machine Translation: Shared Task Papers}, pages 946--954. Association for Computational Linguistics.

\bibitem[{Schwenk et~al.(2021)Schwenk, Wenzek, Edunov, Grave, Joulin, and Fan}]{schwenk-etal-2021-ccmatrix}
Holger Schwenk, Guillaume Wenzek, Sergey Edunov, Edouard Grave, Armand Joulin, and Angela Fan. 2021.
\newblock \href {https://doi.org/10.18653/v1/2021.acl-long.507} {{CCM}atrix: Mining billions of high-quality parallel sentences on the web}.
\newblock In \emph{Proceedings of the 59th Annual Meeting of the Association for Computational Linguistics and the 11th International Joint Conference on Natural Language Processing (Volume 1: Long Papers)}, pages 6490--6500. Association for Computational Linguistics.

\bibitem[{Sloto et~al.(2023)Sloto, Thompson, Khayrallah, Domhan, Gowda, and Koehn}]{sloto-etal-2023-findings}
Steve Sloto, Brian Thompson, Huda Khayrallah, Tobias Domhan, Thamme Gowda, and Philipp Koehn. 2023.
\newblock \href {https://doi.org/10.18653/v1/2023.wmt-1.5} {Findings of the {WMT} 2023 shared task on parallel data curation}.
\newblock In \emph{Proceedings of the Eighth Conference on Machine Translation}, pages 95--102. Association for Computational Linguistics.

\bibitem[{Steingr{\'i}msson et~al.(2023)Steingr{\'i}msson, Loftsson, and Way}]{steingrimsson-etal-2023-filtering}
Stein{\th}{\'o}r Steingr{\'i}msson, Hrafn Loftsson, and Andy Way. 2023.
\newblock \href {https://aclanthology.org/2023.nodalida-1.58/} {Filtering matters: Experiments in filtering training sets for machine translation}.
\newblock In \emph{Proceedings of the 24th Nordic Conference on Computational Linguistics (NoDaLiDa)}, pages 588--600. University of Tartu Library.

\bibitem[{Steingrimsson(2023)}]{steingrimsson-2023-sentence}
Steinthor Steingrimsson. 2023.
\newblock \href {https://doi.org/10.18653/v1/2023.wmt-1.38} {A sentence alignment approach to document alignment and multi-faceted filtering for curating parallel sentence pairs from web-crawled data}.
\newblock In \emph{Proceedings of the Eighth Conference on Machine Translation}, pages 366--374. Association for Computational Linguistics.

\bibitem[{Tiedemann and Thottingal(2020)}]{tiedemann-thottingal-2020-opus}
J{\"o}rg Tiedemann and Santhosh Thottingal. 2020.
\newblock \href {https://aclanthology.org/2020.eamt-1.61/} {{OPUS}-{MT} {--} building open translation services for the world}.
\newblock In \emph{Proceedings of the 22nd Annual Conference of the European Association for Machine Translation}, pages 479--480. European Association for Machine Translation.

\bibitem[{Velayuthan et~al.(2024)Velayuthan, Jayakody, De~Silva, Fernando, and Ranathunga}]{velayuthan-etal-2024-back}
Menan Velayuthan, Dilith Jayakody, Nisansa De~Silva, Aloka Fernando, and Surangika Ranathunga. 2024.
\newblock \href {https://doi.org/10.18653/v1/2024.wmt-1.87} {Back to the stats: Rescuing low resource neural machine translation with statistical methods}.
\newblock In \emph{Proceedings of the Ninth Conference on Machine Translation}, pages 901--907. Association for Computational Linguistics.

\bibitem[{Wijeratne et~al.(2019)Wijeratne, de~Silva, and Shanmugarasa}]{wijeratne2019natural}
Yudhanjaya Wijeratne, Nisansa de~Silva, and Yashothara Shanmugarasa. 2019.
\newblock Natural language processing for government: Problems and potential.
\newblock Technical report, LIRNEasia.

\bibitem[{Zhang et~al.(2020)Zhang, Nagesh, and Knight}]{zhang-etal-2020-parallel-corpus}
Boliang Zhang, Ajay Nagesh, and Kevin Knight. 2020.
\newblock \href {https://doi.org/10.18653/v1/2020.acl-main.756} {Parallel corpus filtering via pre-trained language models}.
\newblock In \emph{Proceedings of the 58th Annual Meeting of the Association for Computational Linguistics}, pages 8545--8554. Association for Computational Linguistics.

\end{thebibliography}

\appendix

\section{Human Evaluation}\label{sec:AppendixHumanEvaluation}

\textbf{Selection of the Annotators: } We select annotators for this task who have translation or machine translation-related experience for a minimum of 2 years and are fluent in both languages of the specific language pair assigned to them. Table~\ref{tab:he_ann_details} shows the years of experience and the qualifications of those annotators who conducted this task.  

\begin{table}[!htp]\centering
\scriptsize
\resizebox{\linewidth}{!}{
\begin{tabular}{lrl}\toprule
\textbf{Annotator} &\textbf{Experience} &\textbf{Qualification} \\
&\textbf{(Years)}&\\
\midrule
Annotator 01 &22 &Diploma in Translation And Interpretation \\
Annotator 02 &9 &BSc.(Hons) in Information Technology \\
Annotator 03 &5 &MBBS \\
Annotator 04 &4	&BA (Hons) in Translation\\
Annotator 05 &3 &BSc (Hons) Engineering sp. in Computer Science and Engineering \\
Annotator 06 & 2.5 &Diploma in Translation And Interpretation\\
Annotator 07 &2.5 &BSc Eng (Hons) Electrical \& Electronics Engineering \\
Annotator 08 &2.5 &BSc (Hons) Engineering sp. in Electrical Engineering \\
Annotator 09 &2 &Bachelor of Industrial Information Technology \\
\bottomrule
\end{tabular}}
\vspace{1mm}
\caption{Annotator details with the years of experience and their qualifications.}\label{tab:he_ann_details}
\end{table}

\noindent\textbf{Resources Provided and Training:} All the annotators have had prior experience with a similar task~\cite{ranathunga-etal-2024-quality}. However, for this annotation work, we provide them with the definitions of the noise categories (Table~\ref{tab:HE_error_taxonomy}) along with example sentence pairs (Table~\ref{tab:examples_noisy_parallel_sentences}) and the guideline in terms of a flowchart (Figure~\ref{fig:human_annotation_flowchart}). First, we asked them to do a sample of 30 sentences as a training on the task, and review it. Then, the 1200 sentence pairs to be annotated were shared with each annotator via Google Sheets to be completed.

\begin{table}[h]
\centering
\scriptsize
\resizebox{\linewidth}{!}{
\renewcommand{\arraystretch}{1.2}
\begin{tabular}{p{0.5\textwidth}} 
\toprule
\textbf{Parallel Corpus Categorization Code: Description} \\ 
\toprule
\textbf{CC}: \textbf{Perfect Translation-pair} \\
Source and target sentences are translation pairs of each other. \\
\hline
\textbf{CN}: \textbf{Near Perfect Translation-pair} \\
Perfect translation pairs. Just a few spelling, grammar, punctuation, or unnecessary characters have to be handled. \\
\hline
\textbf{CB}: \textbf{Low-quality Translation-pair} \\
A full sentence or phrase, but a low-quality (boilerplate) translation. Includes under/over translations. \\
\hline
\textbf{CS}: \textbf{Short Translation Content} \\
Less than 5 words. Translation-wise, correct, but only a short phrase or a few words. \\
\hline
\textbf{CCN}: \textbf{High-overlapping non-translatable text} \\
Perfect or near-perfect translation pair, but with overlapping content like numbers, acronyms, or URLs. Sentences longer than 5 words with high overlap. \\
\hline
\textbf{X}: \textbf{Wrong Translation} \\
Source and target sentences are in the correct languages, but semantically unrelated. Not true translations. \\
\hline
\textbf{UN}: \textbf{Untranslated Text} \\
The source or target is copied from its counterpart (partial or full). Overlapping untranslated content exceeds 30\%. It could have been translated/transliterated. \\
\hline
\textbf{NL}: \textbf{Not a Language} \\
At least one side is not linguistic content. \\
\hline
\textbf{WL}: \textbf{Wrong Language} \\
Either the source or the target (or both) is not in the expected language. Up to 30\% of acceptable content may be tolerated. \\
\bottomrule
\end{tabular}}
\vspace{1mm}
\caption{Improved ~\citet{ranathunga-etal-2024-quality}'s taxonomy to categorize parallel sentences to identify biases in multiPLMs.}
\label{tab:HE_error_taxonomy}
\end{table}

\begin{table*}[!ht]
\centering
\begin{tabular}{c}
\includegraphics[width=0.95\linewidth]{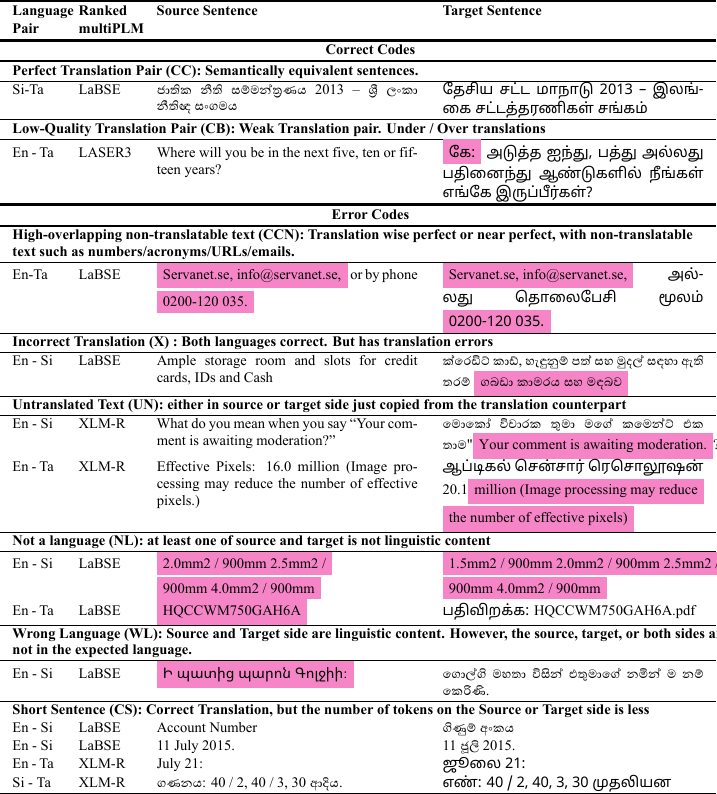}\\
\end{tabular}
\caption{Example parallel sentences from the En-Si, En-Ta and Si-Ta, identified during human evaluation. The translation error in the language pair is highlighted in pink.}
\label{tab:examples_noisy_parallel_sentences}
\end{table*}

\begin{figure*}[!htp]%
\centering
\includegraphics[trim={0cm 0cm 0cm 0cm}, clip, width=0.98\textwidth]{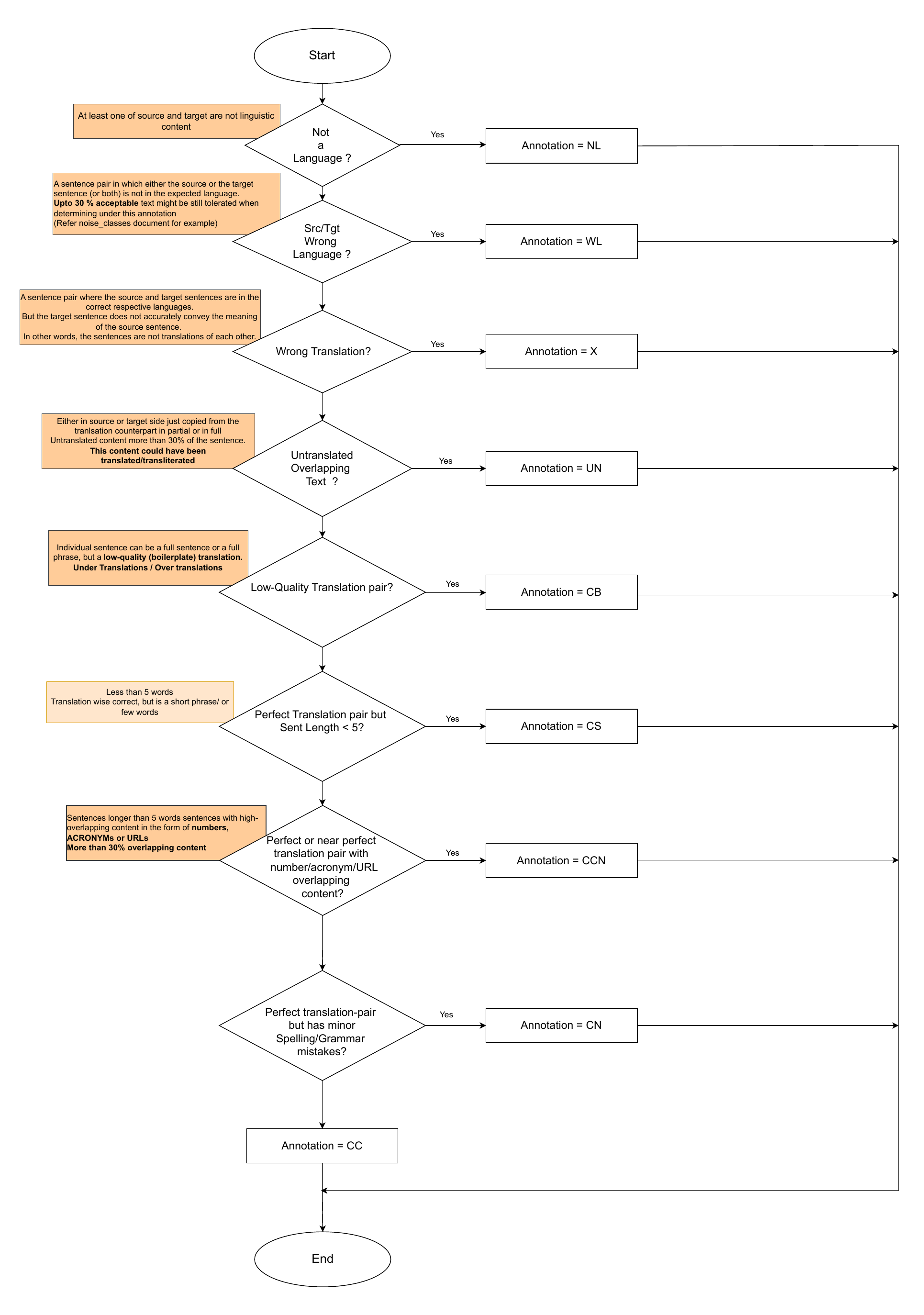}
\caption{Shows the annotation guideline document in terms of a flow chart. This shows the priority of the noise category to be selected prior to declaring the annotation class.}\label{fig:human_annotation_flowchart}
\end{figure*}

\noindent\textbf{Compensation: } They were paid the standard rate in Sri Lanka for each sentence pair they annotated or were offered co-authorship of this paper.

\begin{table}[!ht]
\centering
\renewcommand{\arraystretch}{1.2}
\begin{tabular}{c}
\includegraphics[width=0.9\linewidth]{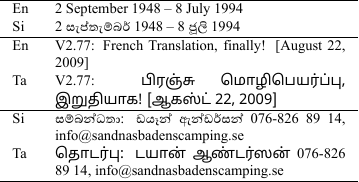} \\
\end{tabular}
\caption{Example parallel sentences which will be separately identified under the new noise category \textit{CCN}}
\label{tab:ccn_example}
\end{table}

\section{Improved Noise Taxonomy}\label{sec:appendixImprovedNoiseTaxonomy}
In this research, we extend the taxonomy of ~\citet{ranathunga-etal-2024-quality} by introducing the noise category \textbf{CCN} (high-overlapping non-translatable text) to capture the type of noise that multiPLMs are biased to rank highly. Table~\ref{tab:ccn_example} shows examples for this noise type. The definitions of all the categories in our improved taxonomy are given in Table~\ref{tab:HE_error_taxonomy}. Then, in Table~\ref{tab:examples_noisy_parallel_sentences}, we show examples of parallel sentences which fall under each of these categories.

\section{Sentence Length Threshold}\label{sec:AppendixSentLengthThresh}
To determine the optimal threshold for sentence length filtering, we filter sentences with fewer than 3, 4, and 5 words on the source (S) side, target (T) side, and both sides. Similar to other PDC experiments, we then rank the parallel sentence pairs by computing embeddings using XLM-R, LABSE and LASER. Finally, we train NMT models using the top 100k-ranked data. We conduct these experiments using the CCAligned corpus for En-Si and En-Ta language pairs. The ChrF++ results are shown in Table~\ref{tab:sLength_threshold}. We observe sentence length threshold of 5 yields the best results consistently. Therefore, we select this threshold in all subsequent experiments. 

\begin{table}[!htp]\centering
\scriptsize
\renewcommand{\arraystretch}{1.2}
\resizebox{1.0\linewidth}{!}{%
\begin{tabular}{llrrrr}\toprule
&\textbf{Side} &\textbf{LASER3} &\textbf{XLM-R} &\textbf{LaBSE} \\\midrule
\multicolumn{5}{c}{\textbf{English - Sinhala}} \\
\hline
Baseline & &32.33 &19.39 &27.57 \\
\hline
Sentence Length (flt < 5) &S &33.86 &26.53 &32.97 \\
&\textbf{T} &\textbf{34.88} &29.42 &33.14 \\
&\textbf{ST} &34.83 &\textbf{29.55} &\textbf{33.50} \\
\hline
Sentence Length (flt < 4) &S &34.25 &23.20 &31.07 \\
&T &33.99 &23.06 &31.30 \\
&ST &34.69 &25.64 &31.55 \\
\hline
Sentence Length (flt < 3) &S &34.50 &25.94 &31.91 \\
&T &34.14 &26.55 &32.49 \\
&ST &34.06 &27.16 &32.76 \\
\hline
\multicolumn{5}{c}{\textbf{English - Sinhala}} \\
\hline
Baseline & &40.13 &17.40 &26.00 \\
\hline
Sentence Length (flt < 5) &S &41.40 &27.60 &36.77 \\
&\textbf{T} &\textbf{41.54} &30.16 &37.61 \\
&\textbf{ST} &41.14 &\textbf{32.67} &\textbf{38.08} \\
\hline
Sentence Length (flt < 4) &S &40.85 &24.28 &36.03 \\
&T &41.15 &26.28 &37.05 \\
&ST &41.52 &30.40 &37.75 \\
\hline
Sentence Length (flt < 3) &S &40.90 &22.13 &33.56 \\
&T &41.42 &24.37 &36.22 \\
&ST &40.58 &25.45 &36.60 \\
\bottomrule
\end{tabular}}
\caption{NMT results in ChrF++ for different sentence length thresholds.}\label{tab:sLength_threshold}
\end{table}

\section{Selection of Languages and Datasets}\label{sec:appendixLangsDatasets}

This section contains details on the selected languages and the web-mined corpora considered under the study.

\paragraph{\textbf{Sinhala}} is an Indo-Aryan language spoken primarily in Sri Lanka by the Sinhalese majority. It exhibits complex morphological structures, including rich inflectional and derivational processes, but is classified as a low-resource language due to the scarcity of linguistic resources and tools~\cite{de2025survey,ranathunga-de-silva-2022-languages}.

\paragraph{\textbf{Tamil},} a Dravidian language with a rich literary history, is spoken by Tamil communities in Sri Lanka, India, and the global diaspora. Unlike Sinhala, Tamil benefits from a relatively larger digital presence, but it still faces challenges in NLP applications due to morphological complexity, agglutinative grammar, and resource limitations in certain domains~\cite{wijeratne2019natural}.

We obtain the publicly released CCMatrix and CCAligned datasets from the OPUS collection~\cite{tiedemann-thottingal-2020-opus}\footnote{https://opus.nlpl.eu/}. Both these datasets support the language pairs, En-Si, En-Ta and Si-Ta, which we consider in this research.
\vspace{10pt}
\newline\textbf{CCMatrix}~\cite{schwenk-etal-2021-ccmatrix} is a web-mined parallel corpus extracted using LASER2-based sentence embeddings to align bitext. While it provides large-scale data, it is highly noisy due to the \textit{global mining} approach to determine alignments, resulting in misaligned or low-quality translations.
\vspace{5pt}
\newline \textbf{CCAligned}~\cite{el-kishky-etal-2020-ccaligned}
extracts bitext from Common Crawl\footnote{https://commoncrawl.org/} using document-level and sentence-level alignment based on multilingual embeddings. Though it improves alignment quality over global bitext-mined corpora, it still contains significant noise, requiring careful filtering for reliable use.

\section{Selection of multiPLMs}\label{sec:AppendixCMultiPLMs}

We include the details on the three multiPLMs considered in this study.

\noindent\textbf{LASER3}~\cite{heffernan-etal-2022-bitext} (L=12, H=1024, A=4, P=250M)\footnote{No. of Layers, Hidden Layer Dimensions, No of Attention Heads, and Total number of parameters are defined by L, H, A, and P respectively.} is a multiPLM favourable for bitext mining and cross-lingual tasks. It improves over previous LASER2 versions by supporting more languages and enhancing alignment quality, but still faces challenges in low-resource settings.

\noindent\textbf{XLM-R}~\cite{conneau-etal-2020-unsupervised} (L=12, H=768, A=6, P=278M) is a transformer-based multiPLM trained on massive amounts of text using masked language modelling. It achieves strong cross-lingual performance but struggles with low-resource languages due to limited training data.

\noindent\textbf{LaBSE}~\cite{feng-etal-2022-language} (L=12, H=768, A=12, P=471M)  is a BERT-based model optimised for multilingual sentence embeddings and bitext retrieval. It provides high-quality cross-lingual representations and is favourable for cross-lingual tasks.

\section{NMT Experiments}\label{sec:AppendixNMTExperiments}
The experiments are conducted on a NVIDIA Quadro RTX6000 GPU with 24GB VRAM. The hyperparameters used during training, along with the training parameters, are shown in Table~\ref{tab:nmt_hyperparameters}. We conduct training on the NMT systems for 100 epochs with early stopping criteria and report the results using ChrF++. ChrF++ was chosen over the conventional multi-BLEU~\cite{papineni-etal-2002-bleu} and sacreBLEU~\citep{post-2018-call} because character-level evaluation is more suitable for the considered languages, Sinhala and Tamil, which are morphologically rich in nature.

\begin{table}[!htp]\centering
\scriptsize
\begin{tabular*}{0.4\textwidth}{@{\extracolsep\fill}lr}
\toprule
\textbf{Hyperparameter} &\textbf{Argument value} \\
\midrule
encoder/decoder Layers &6 \\
encoder/decoder attention heads &4\\
encoder-embed-dim &512\\
decoder-embed-dim &512\\
encoder-ffn-embed-dim &2048 \\
decoder-ffn-embed-dim &2048 \\
dropout &0.4\\
attention-dropout &0.2\\
optimizer &adam\\
Adam $\beta_1$, Adam $\beta_2$ &0.9, 0.99 \\
warmup-updates &4000\\
warmup-init-lr &1e-7\\
learning rate &1e-3\\
batch-size &32\\
patience &6\\
fp16 &True\\
\midrule
\end{tabular*}
\caption{Training parameters for NMT experiments.}\label{tab:nmt_hyperparameters}
\end{table}

\begin{table*}[!h]
\centering
\small
\renewcommand{\arraystretch}{1.1}
\resizebox{\textwidth}{!}{%
\begin{tabular}{lrrrlrrr}
\noalign{\hrule height 1.2pt}
\multirow{2}{*}{\textbf{Heustic/s}} &\multicolumn{3}{c}{\textbf{CCMatrix}} &\multirow{2}{*}{\textbf{Heustic/s}} &\multicolumn{3}{c}{\textbf{CCAligned}} \\\cmidrule{2-4}\cmidrule{6-8}
&\textbf{LASER3} &\textbf{XLM-R} &\textbf{LaBSE} & &\textbf{LASER3} &\textbf{XLM-R} &\textbf{LaBSE} \\
\noalign{\hrule height 1.2pt}
\multicolumn{8}{c}{\textbf{Sinhala →Tamil}} \\
\noalign{\hrule height 1.2pt}
Baseline &31.08 &30.99 &31.63 &Baseline &35.36 &35.97 &35.79 \\
\noalign{\hrule height 0.8pt}
\multicolumn{8}{l}{\textbf{Deduplication-based Heuristics}} \\
DD+PN+5gram (ST+T) &\colorbox{\cGreen}{\ul{\textbf{32.98}}} &\colorbox{\cGreen}{\ul{\textbf{32.73}}} &\colorbox{\cGreen}{\ul{\textbf{32.60}}} &DD+puntsNums (T) &36.63 &36.47 &\ul{\textbf{36.86}} \\
&\textbf{} &\textbf{} &\textbf{} &DD+PN (ST) &35.96 &\textbf{36.71} &36.23 \\
\textbf{} & & & &DD+PN+7gram &\ul{\textbf{36.73}} &36.62 &36.37 \\
\multicolumn{8}{l}{\textbf{Length-based Heurics}} \\
 (S) &31.41 &31.52 &32.30 & (T) &36.30 &\textbf{36.71} &36.58 \\
& & & & (ST) &\textbf{36.47} &35.99 &\textbf{36.60} \\
\multicolumn{8}{l}{\textbf{LID-based Heuristics}} \\
LID (S) &31.48 &31.36 &31.78 &LIDThresh (ST) &\ul{\textbf{36.73}} &\ul{\textbf{36.73}} &36.80 \\
\multicolumn{8}{l}{\textbf{Ratio-based Heuristics}} \\
sentCRatio (ST) &32.28 &31.90 &32.04 &sentWRatio (T) &36.24 &36.17 &\textbf{36.46} \\
& & & &sentWRatio (ST) &\textbf{36.44} &\textbf{36.72} &36.01 \\
\multicolumn{8}{l}{\textbf{Combined Heuristics}} \\
\noalign{\hrule height 0.8pt}
DD+PN+5gram++sentCharRatio &31.45 &\textbf{32.65} &31.17 &DD+PN+7gram++sentWRatio &36.60 &\colorbox{\cGreen}{\textbf{36.85}} &36.32 \\
DD+PN+5gram++LIDThresh+sentCharRatio &\textbf{32.64} &31.30 &\textbf{32.28} &DD+PN+7gram++LIDThresh+sentWRatio0.8 &\colorbox{\cGreen}{\textbf{36.83}} &36.66 &\colorbox{\cGreen}{\textbf{37.03}} \\
\noalign{\hrule height 0.8pt}
Gain (DD-heuristic vs Baseline) &1.90 &1.74 &0.97 & &1.37 &0.50 &1.07 \\
Gain (sLength vs Baseline) &0.33 &0.53 &0.67 & &0.94 &0.74 &0.79 \\
Gain (LID-heuristic vs Baseline) &0.40 &0.37 &0.15 & &1.37 &0.76 &1.01 \\
Gain (Raio heuristic vs Baseline) &1.20 &0.91 &0.41 & &1.08 &0.75 &0.67 \\
\noalign{\hrule height 0.8pt}
\textbf{Gain (Overall individual Heuristic vs Baseline)} &\textbf{1.90} &\textbf{1.74} &\textbf{0.97} &\textbf{} &\textbf{1.37} &\textbf{0.76} &\textbf{1.07} \\
\textbf{Gain (Combined Heuristic vs Baseline)} &\textbf{1.56} &\textbf{1.66} &\textbf{0.65} &\textbf{} &\textbf{1.47} &\textbf{0.88} &\textbf{1.24} \\
\textbf{Gain (Combined vs Individual )} &\ul{\textbf{-0.34}} &\ul{\textbf{-0.08}} &\ul{\textbf{-0.32}} &\textbf{} &\ul{\textbf{0.10}} &\ul{\textbf{0.12}} &\ul{\textbf{0.17}} \\
\noalign{\hrule height 1.2pt}
\multicolumn{8}{c}{\textbf{English →Sinhala}} \\
\noalign{\hrule height 1.2pt}
Baseline &30.76 &5.55 &14.49 &Baseline &32.33 &19.39 &27.57 \\
\multicolumn{8}{l}{\textbf{Deduplication-based Heuristics}} \\
DD+puntsNums (T) &33.89 &14.81 &\textbf{26.31} &DD+puntsNums+5gram (ST+T) &33.81 &\textbf{30.33} &\textbf{32.74} \\
DD+puntsNums+5gram (T+T) &\ul{\textbf{34.50}} &\ul{\textbf{16.09}} &25.78 &DD+puntsNums+6gram (ST+T) &\textbf{35.24} &28.21 &31.26 \\
\multicolumn{8}{l}{\textbf{Length-based Heurics}} \\
sLength (ST) &32.82 &8.24 &\ul{\textbf{29.96}} &sLength (T) &\textbf{34.88} &29.42 &33.14 \\
& & &\textbf{} &sLength (ST) &34.83 &\textbf{29.55} &\ul{\textbf{33.50}} \\
\multicolumn{8}{l}{\textbf{LID-based Heuristics}} \\
LID &31.99 &13.32 &\textbf{16.20} &LIDThresh (S) &\ul{\textbf{35.73}} &30.86 &32.69 \\
LIDThresh &\textbf{32.84} &\textbf{14.08} &13.71 &LIDThresh (ST) &35.11 &\ul{\textbf{32.97}} &\textbf{32.88} \\
\multicolumn{8}{l}{\textbf{Ratio-based Heuristics}} \\
sentWRatio (S) &\textbf{31.50} &\textbf{7.40} &10.86 &sentWRatio (S) &\textbf{34.15} &25.97 &\textbf{31.35} \\
sentWRatio (ST) &30.64 &7.00 &\textbf{15.50} &sentWRatio (ST) &33.85 &\textbf{28.73} &31.17 \\
\multicolumn{8}{l}{\textbf{Combined Heuristics}} \\
\noalign{\hrule height 0.5pt}
DD+PN+5gram +sLength +LIDThresh +sentWRatio &\colorbox{\cGreen}{\textbf{36.10}} &23.84 &\colorbox{\cGreen}{\textbf{33.94}} &DD+PN+5gram+sLength+LIDThresh+sentWRatio &36.15 &34.50 &\colorbox{\cGreen}{\textbf{35.67}} \\
DD+PN+5gram+sLength+LIDThresh+sentWRatio>0.8 &35.66 &\colorbox{\cGreen}{\textbf{24.18}} &33.19 &DD+PN+5gram+sLength+LIDThresh+sentWRatio>0.8 &\colorbox{\cGreen}{\textbf{36.26}} &\colorbox{\cGreen}{\textbf{35.66}} &35.42 \\
\noalign{\hrule height 0.8pt}
Gain (Dedup-heuristic vs Baseline) &3.74 &10.54 &11.82 & &2.91 &10.94 &5.17 \\
Gain (sLength vs Baseline) &2.06 &2.69 &15.47 & &2.55 &10.16 &5.93 \\
Gain (LID-heuristic vs Baseline) &2.08 &8.53 &1.71 & &3.40 &13.58 &5.31 \\
Gain (Raio heuristic vs Baseline) &0.74 &1.85 &1.01 & &1.82 &9.34 &3.78 \\
\noalign{\hrule height 0.8pt}
\textbf{Gain (Overall individual Heuristic vs Baseline)} &\textbf{3.74} &\textbf{10.54} &\textbf{15.47} &\textbf{} &\textbf{3.40} &\textbf{13.58} &\textbf{5.93} \\
\textbf{Gain (Combined Heuristic vs Baseline)} &\textbf{5.34} &\textbf{18.63} &\textbf{19.45} &\textbf{} &\textbf{3.93} &\textbf{16.27} &\textbf{8.10} \\
\textbf{Gain (Combined vs Individual )} &\ul{\textbf{1.60}} &\ul{\textbf{8.09}} &\ul{\textbf{3.98}} &\textbf{} &\ul{\textbf{0.53}} &\ul{\textbf{2.69}} &\ul{\textbf{2.17}} \\
\noalign{\hrule height 1.2pt}
\multicolumn{8}{c}{\textbf{English →Tamil}} \\
\noalign{\hrule height 1.2pt}
Baseline &19.02 &5.86 &14.20 &Baseline &40.13 &17.40 &26.00 \\
\multicolumn{8}{l}{\textbf{Deduplication-based Heuristics}} \\
DD-7gram (T) &18.15 &7.37 &\textbf{21.96} &DD+puntsNums+4gram (ST+T) &41.82 &35.90 &\textbf{37.08} \\
DD+puntNums (T) &\textbf{21.57} &\textbf{8.24} &20.41 &DD+puntsNums+6gram (ST+ST) &\textbf{41.90} &\textbf{35.97} &35.94 \\
\multicolumn{8}{l}{\textbf{Length-based Heurics}} \\
sLength (T) &18.52 &\textbf{6.33} &\textbf{21.73} &sLength (T) &\textbf{41.54} &30.16 &37.61 \\
sLength (ST) &\textbf{19.45} &5.33 &20.79 &sLength (ST) &41.14 &\textbf{32.67} &\ul{\textbf{38.08}} \\
\multicolumn{8}{l}{\textbf{LID-based Heuristics}} \\
LIDThresh (T) &\ul{\textbf{29.59}} &\ul{\textbf{15.24}} &24.51 &LIDThresh ST) &\ul{\textbf{42.63}} &\ul{\textbf{38.01}} &\textbf{37.40} \\
LIDThresh (ST) &28.93 &15.16 &\ul{\textbf{25.33}} & &\textbf{} &\textbf{} & \\
\multicolumn{8}{l}{\textbf{Ratio-based Heuristics}} \\
STRatio &\textbf{20.52} &5.40 &\textbf{18.29} &sentWRatio (S) &\textbf{42.05} &29.70 &35.53 \\
sentCRatio (T) &19.90 &\textbf{6.78} &12.51 &sentWRatio (ST) &41.05 &\textbf{30.88} &\textbf{35.77} \\
\multicolumn{8}{l}{\textbf{Combined Heuristics}} \\
\noalign{\hrule height 0.8pt}
DD+PN+7gram+slength+LIDThresh+STRatio &\colorbox{\cGreen}{\textbf{30.67}} &\colorbox{\cGreen}{\textbf{23.36}} &\colorbox{\cGreen}{\textbf{31.80}} &DD+PN+6gram+sLength+LIDThresh+sentWRatio &\colorbox{\cGreen}{\textbf{43.47}} &\colorbox{\cGreen}{\textbf{41.74}} &41.06 \\
& & & &DD+PN+6gram+sLength+LIDThresh+sentWRatio > 0.8 &42.08 &40.56 &\colorbox{\cGreen}{\textbf{42.02}} \\
\noalign{\hrule height 0.8pt}
Gain (Dedup-heuristic vs Baseline) &2.55 &2.38 &7.76 & &1.77 &18.57 &11.08 \\
Gain (sLength vs Baseline) &0.43 &0.47 &7.53 & &1.41 &15.27 &12.08 \\
Gain (LID-heuristic vs Baseline) &10.57 &0.92 &10.31 & &2.50 &20.61 &11.40 \\
Gain (Raio heuristic vs Baseline) &1.50 &0.92 &4.09 & &1.92 &13.48 &9.77 \\
\noalign{\hrule height 0.8pt}
\textbf{Gain (Overall individual Heuristic vs Baseline)} &\textbf{10.57} &\textbf{2.38} &\textbf{10.31} &\textbf{} &\textbf{2.50} &\textbf{20.61} &\textbf{12.08} \\
\textbf{Gain (Combined Heuristic vs Baseline)} &\textbf{11.65} &\textbf{17.50} &\textbf{17.60} &\textbf{} &\textbf{1.92} &\textbf{12.30} &\textbf{16.02} \\
\textbf{Gain (Combined vs Individual )} &\ul{\textbf{1.08}} &\ul{\textbf{8.12}} &\ul{\textbf{6.47}} &\textbf{} &\ul{\textbf{0.84}} &\ul{\textbf{3.73}} &\ul{\textbf{3.94}} \\
\noalign{\hrule height 1.2pt}
\end{tabular}}
\caption{Shows the \textbf{best} NMT performance produced using the individual heuristics as well as combinations of heuristics. The values in bold indicate the highest NMT score obtained for a given heuristic or heuristic combination. The values underlined are the highest among the individual heuristics. Highlighted in green are the overall best values. Here \textbf{DD+PN} is \textit{Deduplication+punctNums}, \textbf{SL} is \textit{sLength} and \textbf{LT} is \textit{LIDThresh}.}\label{tab:nmt_best_summary}
\end{table*}

\begin{figure}[!t]%
\centering
\includegraphics[trim={0cm 0cm 0cm 0cm}, clip, width=0.85\linewidth]{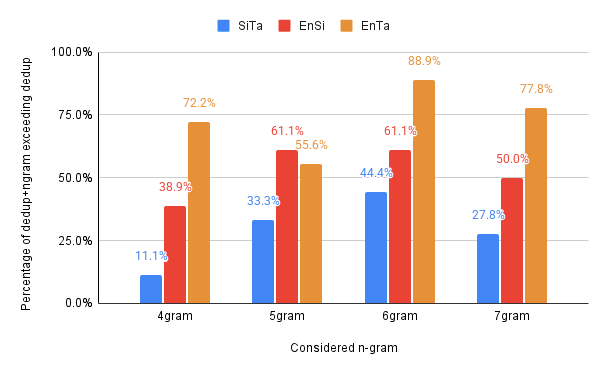}
\caption{Percentage of \textit{dedup+ngram} experiments exceeding the best result of \textit{dedup} with respect to the {Language-pair}. }\label{fig:ngram_vs_langpair}
\end{figure}

\begin{figure*}[!htb]
    \centering
    \begin{subfigure}{0.32\textwidth}  \includegraphics[width=\linewidth]{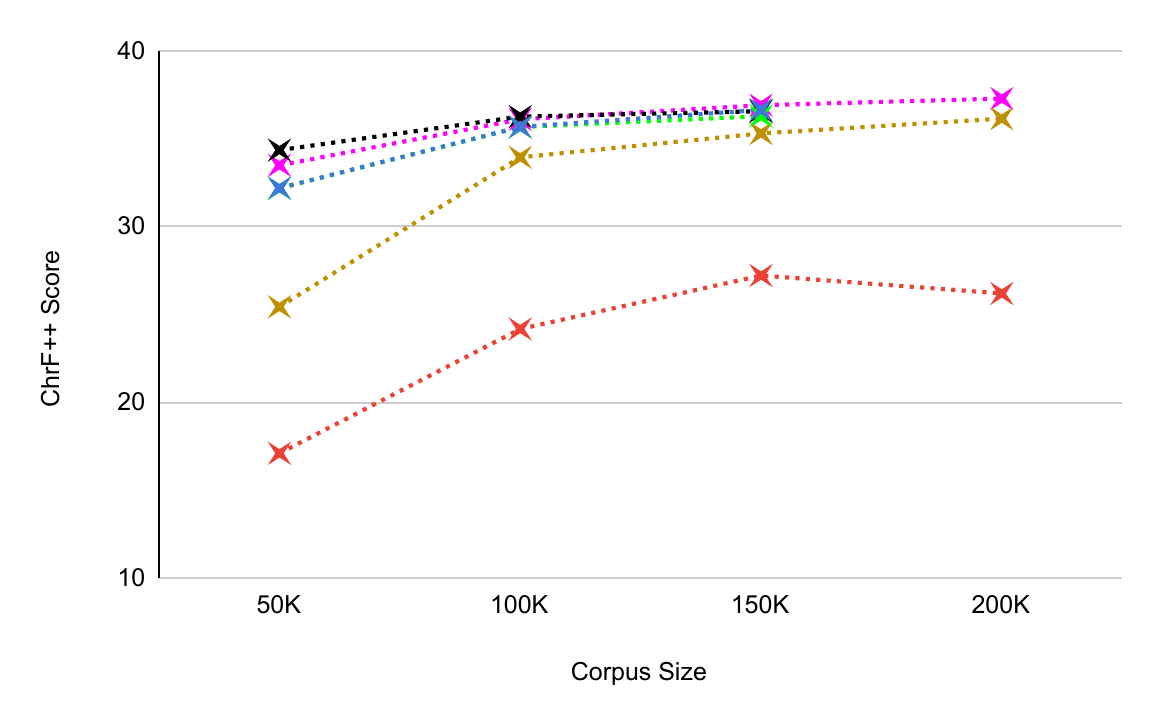}
        \caption{\centering En-Si}
        \label{fig:imageensi}
    \end{subfigure}
    \begin{subfigure}{0.32\textwidth}   \includegraphics[width=\linewidth]{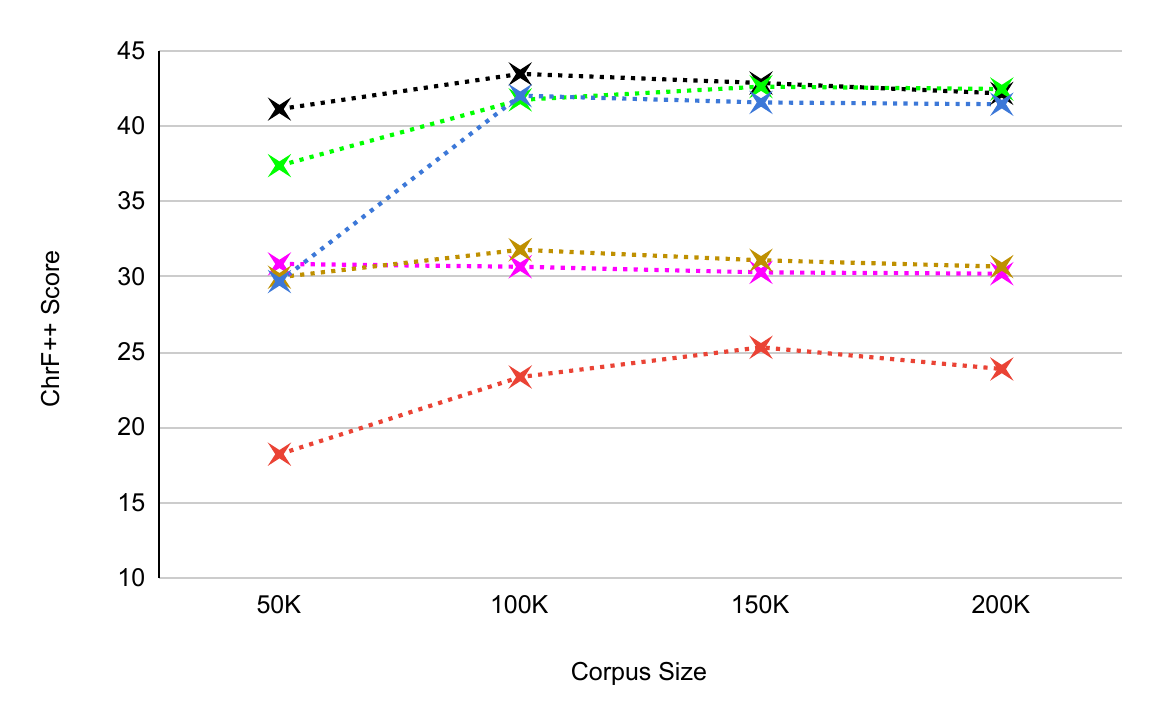}
        \caption{\centering En-Ta}
        \label{fig:imageenta}
    \end{subfigure}
    \begin{subfigure}{0.32\textwidth}  \includegraphics[width=\linewidth]{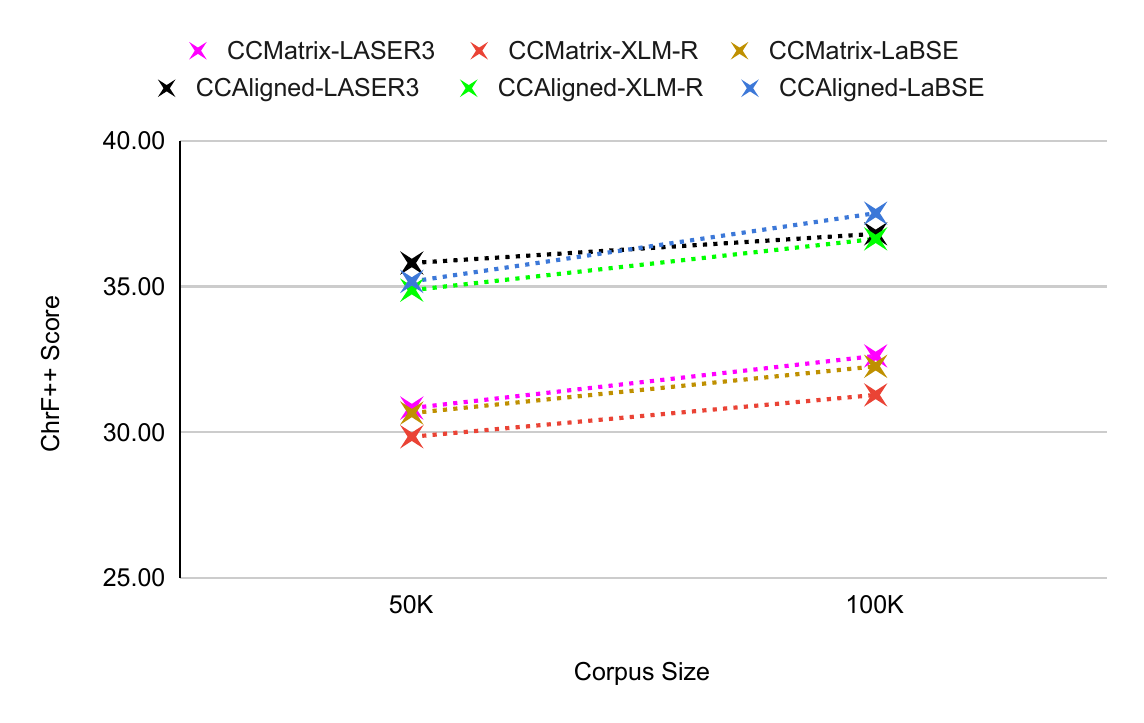}
        \caption{\centering Si-Ta}
        \label{fig:imagesita}
    \end{subfigure}  
    \caption{ChrF++ scores of NMT systems trained by varying the training dataset size.}\label{fig:size_vs_nmt}
\end{figure*}

\section{Results Analysis}\label{sec:AppendixResultsAnaysis}

We show the individual and the combined heuristic which produced the best NMT gains with respect to each multiPLM, dataset and the language pair in the Table~\ref{tab:nmt_best_summary}. 
The Figure~\ref{fig:ngram_vs_langpair} shows the percentage of  \textit{dedup+n-gram} experiments exceeding the highest \textit{dedup}, with respect to the language-pair.
Finally, we show the final dataset sizes after applying the heuristic along with the percentage reduction in Table~\ref{tab:af_dataset_sizes}.

\begin{table*}[h]\centering
\scriptsize
\renewcommand{\arraystretch}{1.3}
\resizebox{1.0\textwidth}{!}{%
\begin{tabular}{llrrrrrrrrrrrrr}
\toprule
\multirow{3}{*}{\textbf{Heuristic}} &\multirow{3}{*}{\textbf{Applicable Side}} &\multicolumn{4}{c}{\textbf{Sinhala - Tamil}} &\multicolumn{4}{c}{\textbf{English - Sinhala}} &\multicolumn{4}{c}{\textbf{English - Tamil}} \\\cmidrule{3-14}
& &\multicolumn{2}{c}{\textbf{CCMatrix}} &\multicolumn{2}{c}{\textbf{CCAligned}} &\multicolumn{2}{c}{\textbf{CCMatrix}} &\multicolumn{2}{c}{\textbf{CCAligned}} &\multicolumn{2}{c}{\textbf{CCMatrix}} &\multicolumn{2}{c}{\textbf{CCAligned}} \\
\cmidrule{3-14}
& &\textbf{Dataset Size} &\textbf{\% Reduction} &\textbf{Dataset Size} &\textbf{\% Reduction} &\textbf{Dataset Size} &\textbf{\% Reduction} &\textbf{Dataset Size} &\textbf{\% Reduction} &\textbf{Dataset Size} &\textbf{\% Reduction} &\textbf{Dataset Size} &\textbf{\% Reduction} \\
\midrule
Baseline & &215965 & &260119 & &6270800 & &619730 & &7291118 & &880568 & \\
\hline
DD &S &189654 & &250038 &3.88\% &6146819 &1.98\% &570768 &7.90\% &6378607 &12.52\% &797071 &9.48\% \\
&T &209461 & &247176 &4.98\% &3242950 &48.28\% &562088 &9.30\% &4754106 &34.80\% &780355 &11.38\% \\
&ST &183904 & &243384 &6.43\% &3176145 &49.35\% &537581 &13.26\% &4060447 &44.31\% &736212 &16.39\% \\
DD-4gram &S &176590 &18.23\% &189218 &27.26\% &4171884 &33.47\% &403993 &34.81\% &4802654 &34.13\% &440248 &50.00\% \\
&T &196538 &9.00\% &207603 &20.19\% &2751819 &56.12\% &423752 &31.62\% &4271550 &41.41\% &549621 &37.58\% \\
&ST &172440 &20.15\% &184159 &29.20\% &2035282 &67.54\% &355790 &42.59\% & &100.00\% &365648 &58.48\% \\
DD-5gram &S &188457 &12.74\% &217733 &16.29\% &4486108 &28.46\% &481021 &22.38\% &5104643 &29.99\% &558504 &36.57\% \\
&T &204045 &5.52\% &226738 &12.83\% &3071693 &51.02\% &499307 &19.43\% &4516819 &38.05\% &638600 &27.48\% \\
&ST &185915 &13.91\% &216389 &16.81\% &2374578 &62.13\% &446838 &27.90\% &3319964 &54.47\% &487465 &44.64\% \\
DD-6gram &S &196194 &9.15\% &232604 &10.58\% &5383674 &14.15\% &528525 &14.72\% &5309083 &27.18\% &639961 &27.32\% \\
&T &200310 &7.25\% &237467 &8.71\% &3142124 &49.89\% &539513 &12.94\% &4590987 &37.03\% &681186 &22.64\% \\
&ST &196194 &9.15\% &233792 &10.12\% &2859756 &54.40\% &505429 &18.44\% &3489629 &52.14\% &570403 &35.22\% \\
DD-7gram &S &200899 &6.98\% &240704 &7.46\% &5701801 &9.07\% &554025 &10.60\% &5718913 &21.56\% &679750 &22.81\% \\
&T &204485 &5.32\% &244007 &6.19\% &3170538 &49.44\% &561285 &9.43\% &4631486 &36.48\% &703950 &20.06\% \\
&ST &200898 &6.98\% &246104 &5.39\% &3021457 &51.82\% &538898 &13.04\% &3745771 &48.63\% &611821 &30.52\% \\
DD+N &S &182386 &15.55\% &260119 &0.00\% &6105433 &2.64\% &505828 &18.38\% &6337285 &13.08\% &683882 &22.34\% \\
&T &201551 &6.67\% &218980 &15.82\% &3225979 &48.56\% &502971 &18.84\% &4722644 &35.23\% &675627 &23.27\% \\
&ST &176040 &18.49\% &216238 &16.87\% &3158067 &49.64\% &476379 &23.13\% &4031459 &44.71\% &631485 &28.29\% \\
DD+PN &S &180380 &16.48\% &215100 &17.31\% &5931349 &5.41\% &494778 &20.16\% &6194331 &15.04\% &668849 &24.04\% \\
&T &198352 &8.16\% &212341 &18.37\% &3197186 &49.01\% &492801 &20.48\% &4666158 &36.00\% &660902 &24.95\% \\
&ST &173804 &19.52\% &207197 &20.35\% &3130297 &50.08\% &465617 &24.87\% &3987832 &45.31\% &616702 &29.97\% \\
DD+PN+4gram &ST+T & & & & & & &289248 &53.33\% & & & & \\
DD+PN+5gram &ST + T &167022 &22.66\% &187250 &28.01\% &3044520 &51.45\% &380146 &38.66\% & & & & \\
DD+PN+6gram &ST + T &169784 &21.38\% &189060 &27.32\% & & &428939 &30.79\% &4547759 &37.63\% &464424 &47.26\% \\
DD+PN+7gram &T +T & & &196221 &24.56\% & & & & &4620008 &36.64\% & & \\
\hline
SL &S &150094.00 &30.50\% &188061 &27.70\% &5088747 &18.85\% &411474.00 &33.60\% &6498956 &10.86\% &595057 &32.42\% \\
&T &100799.00 &53.33\% &161363 &37.97\% &3670963 &41.46\% &377708.00 &39.05\% &4267495 &41.47\% &517516 &41.23\% \\
&ST &96264.00 &55.43\% &157978 &39.27\% &3341564 &46.71\% &348829.00 &43.71\% &4134919 &43.29\% &491207 &44.22\% \\
\hline
LID &S &192377.00 &10.92\% &241617 &7.11\% &6200355 &1.12\% &479589.00 &22.61\% &7210848 &1.10\% &669260 &24.00\% \\
&T &186720.00 &13.54\% &241863 &7.02\% &6066681 &3.26\% &575298.00 &7.17\% &6800923 &6.72\% &794143 &9.81\% \\
&ST &178276.00 &17.45\% &231619 &10.96\% &6010065 &4.16\% &457639.00 &26.16\% &6743988 &7.50\% &625281 &28.99\% \\
LT &S &181470.00 &15.97\% &227791 &12.43\% &6120792 &2.39\% &398272.00 &35.73\% &6120793 &16.05\% &564870 &35.85\% \\
&T &172726.00 &20.02\% &222290 &14.54\% &5990169 &4.48\% &546472.00 &11.82\% &5990170 &17.84\% &731484 &16.93\% \\
&ST &162777.00 &24.63\% &208644 &19.79\% &5877142 &6.28\% &377579.00 &39.07\% &5877143 &19.39\% &518010 &41.17\% \\
\hline
STRatio &- &170168.00 &21.21\% &229101 &11.92\% &4293239 &31.54\% &459473.00 &25.86\% &4051888 &44.43\% &679820 &22.80\% \\
sentWRatio &S &199788 &7.49\% &246908 &5.08\% &6232528 &0.61\% &546460 &11.82\% &7231531 &0.82\% &743624 &15.55\% \\
&T &196812 &8.87\% &250151 &3.83\% &6198124 &1.16\% &552798 &10.80\% &7176111 &1.58\% &745221 &15.37\% \\
&ST &193989 &10.18\% &245161 &5.75\% &6177212 &1.49\% &531963 &14.16\% &7138854 &2.09\% &717159 &18.56\% \\
sentCRatio &S &212287 &1.70\% &224031 &13.87\% &6262297 &0.14\% &594169 &4.12\% &7252444 &0.53\% &832611 &5.45\% \\
&T &212877 &1.43\% &218726 &15.91\% &6261991 &0.14\% &596346 &3.77\% &7281050 &0.14\% &851343 &3.32\% \\
&ST &211661 &1.99\% &215151 &17.29\% &6257079 &0.22\% &588310 &5.07\% &7247298 &0.60\% &826866 &6.10\% \\
\hline
\multicolumn{14}{l}{\textbf{Combined Heuristics}}\\
\hline
\multicolumn{14}{l}{DD+PN+ngram (SiTa-CCMatrix n=5, SiTa-CCAligned n= 7 EnSi-CCMatrix/CCAligned n=5, EnTa-CCMatrix n=7, EnTa-CCAligned n=6)} \\
\hline
+SL &T+ST &117198 &45.73\% &143919 &44.67\% &2245307 &64.19\% &240086 &61.26\% &3462458 &52.51\% &321003 &63.55\% \\
+LT &T+ST &130831 &39.42\% &162543 &37.51\% &2880530 &54.06\% &239144 &61.41\% &2876818 &60.54\% &306352 &65.21\% \\
+sentWRatio &T+S &154028 &28.68\% &170715 &34.37\% &2993730 &52.26\% &337634 &45.52\% &3377988 &53.67\% &427587 &51.44\% \\
+SL+LT &T+ ST &99207 &54.06\% &127284 &51.07\% &2200195 &64.91\% &180731 &70.84\% &4330140 &40.61\% &241679 &72.55\% \\
+SL+sentWRatio &T+ST+ST &116344 &46.13\% &127035 &51.16\% &2241513 &64.25\% &224542 &63.77\% &2794637 &61.67\% &298141 &66.14\% \\
+SL+LT+sentWRatio &T+ST+ST+S &127188 &41.11\% &117970 &54.65\% &2197629 &64.95\% &177711 &71.32\% &2726087 &62.61\% &237105 &73.07\% \\
+SL+LT+sentWRatio>0.8 &T+ST+ST+ST &179984 &16.66\% &99311 &61.82\% &2149037 &65.73\% &161869 &73.88\% &2203591 &69.78\% &214639 &75.62\% \\
+SL+LT+sentCRatio &T+ST+ST+ST &98894 &54.21\% &\multicolumn{2}{c}{NA} &\multicolumn{2}{c}{NA} &\multicolumn{2}{c}{NA} &\multicolumn{2}{c}{NA} &\multicolumn{2}{c}{NA} \\
+SL+LT+STRatio &T+ST+ST+STR &82866 &61.63\% &\multicolumn{2}{c}{NA} &\multicolumn{2}{c}{NA} &\multicolumn{2}{c}{NA} &2129744 &70.79\% &\multicolumn{2}{c}{NA} \\
\bottomrule
\end{tabular}}
\caption{Shows the final corpus sizes after applying heuristics along with the reduction percentage. Here \textbf{DD+PN} is \textit{Deduplication+punctNums}, \textbf{SL} is \textit{sLength} and \textbf{LT} is \textit{LIDThresh}. \textbf{NA} corresponds to the experiments that are not applicable for the language-pair.}\label{tab:af_dataset_sizes}
\end{table*}



%

\section{NMT Performance on Training Dataset}\label{sec:AppendixNMTVsTrainingData}

Figure~\ref{fig:size_vs_nmt} shows the variation in NMT results by varying the dataset size with 50k, 100, 150k and 200k for each language-pair.

\end{document}